\documentclass[iicol]{sn-jnl}% Math and Physical Sciences Numbered Reference Style
\usepackage{graphicx}%
\usepackage{multirow}%
\usepackage{amsmath,amssymb,amsfonts}%
\usepackage{amsthm}%
\usepackage{mathrsfs}%
\usepackage[title]{appendix}%
\usepackage{xcolor}%
\usepackage{textcomp}%
\usepackage{manyfoot}%
\usepackage{booktabs}%
\usepackage{algorithm}%
\usepackage{algpseudocode}%
\usepackage{listings}%
\usepackage{array}
\usepackage{natbib} % has a nice set of citation styles and commands
\usepackage{subcaption}

\theoremstyle{thmstyleone}%
\theoremstyle{thmstyletwo}%

\theoremstyle{thmstylethree}%

\begin{document}

\title[Assessing Representation Semantics Under Language Pretraining in Fine-Grained Classification]{Assessing Representation Semantics Under Language Pretraining in Fine-Grained Classification}
% \title{Evaluating Representation Semantics Under Language Pretraining in Fine-Grained Classification}
%%=============================================================%%
%% GivenName	-> \fnm{Joergen W.}
%% Particle	-> \spfx{van der} -> surname prefix
%% FamilyName	-> \sur{Ploeg}
%% Suffix	-> \sfx{IV}
%% \author*[1,2]{\fnm{Joergen W.} \spfx{van der} \sur{Ploeg} 
%%  \sfx{IV}}\email{iauthor@gmail.com}
%%=============================================================%%
\author*[1]{\fnm{Md Raisul} \sur{Kibria}}\email{raisul.kibria@abo.fi}
\author[1]{\fnm{Sébastien} \sur{Lafond}}
% \author[1,2]{Further~Coauthor}
% \author[3]{Further~Coauthor}
% \author[1]{Further~Coauthor}
% \author[3]{Further~Coauthor}
% \author[3,1]{Further~Coauthor}
% Add affiliations after the authors
\affil[1]{%
\orgdiv{Faculty of Science and Engineering}, \orgname{Åbo Akademi University}, \orgaddress{\city{Turku}, \postcode{20500}, \country{Finland}}
}
% \author*[1,2]{\fnm{First} \sur{Author}}\email{iauthor@gmail.com}

% \author[2,3]{\fnm{Second} \sur{Author}}\email{iiauthor@gmail.com}
% \equalcont{These authors contributed equally to this work.}

% \author[1,2]{\fnm{Third} \sur{Author}}\email{iiiauthor@gmail.com}
% \equalcont{These authors contributed equally to this work.}

% \affil*[1]{\orgdiv{Department}, \orgname{Organization}, \orgaddress{\street{Street}, \city{City}, \postcode{100190}, \state{State}, \country{Country}}}

% \affil[2]{\orgdiv{Department}, \orgname{Organization}, \orgaddress{\street{Street}, \city{City}, \postcode{10587}, \state{State}, \country{Country}}}

% \affil[3]{\orgdiv{Department}, \orgname{Organization}, \orgaddress{\street{Street}, \city{City}, \postcode{610101}, \state{State}, \country{Country}}}

%%==================================%%
%% Sample for unstructured abstract %%
%%==================================%%

\abstract{Understanding how supervision signals shape semantic visual representations remains challenging, particularly in fine-grained, low-data settings. Standard evaluations of vision models rely on downstream accuracy, which often fails to capture structural differences in learned representations. We propose a compact explainability-driven framework for comparing encoders trained with classification-only and caption-generation supervision on CUB-200. Using a controlled two-stage protocol in which both encoders are ultimately optimized for the same classification task, we enable a direct comparison of the resulting representations. Saliency maps are generated using attention rollout and gradient-weighted rollout and evaluated with five metrics covering faithfulness (positive and negative perturbation) and localization (keypoint emphasis, background attribution, and Earth Mover's Distance). Across random seeds and explanation methods, caption supervision yields representations that are more sensitive to the removal of salient regions, better aligned with annotated object parts, and less reliant on background regions, whereas classification supervision produces representations that are slightly more robust to the removal of low-saliency regions. An ablation study on the Oxford-IIIT Pet dataset shows consistent trends across key metrics. These results demonstrate that the proposed framework makes supervision-induced representational differences observable and quantifiable beyond downstream accuracy alone. More broadly, it provides a systematic basis for assessing learned representations and comparing alternative training and transfer-learning strategies.}

\keywords{ Vision-Language Supervision, Representation Explainability, Object-Centric Visual Representations, Multitask Representation Transfer.}

%%\pacs[JEL Classification]{D8, H51}

%%\pacs[MSC Classification]{35A01, 65L10, 65L12, 65L20, 65L70}

\maketitle

\section{Introduction}

Cross-modal representation learning has become central to modern machine learning, particularly in vision–language modeling. A core motivation is that natural language supervision can provide semantically rich signals beyond categorical labels. Contrastive pretraining approaches such as in CLIP~\citep{radford2021learning} demonstrate that aligning image and text embeddings at scale yields transferable visual representations. Hybrid objectives further combine contrastive alignment with generative modeling, as in CoCa~\citep{yu2022coca}. Other work studies modality imbalance~\citep{zhang2024multimodal,huang2025adaptive} or integrates contrastive and discriminative objectives within unified architectures~\citep{wei2023iclip}. Despite these advances, it remains challenging to quantify whether different supervision modalities induce systematically different visual representation structures.

Among common visual objectives, classification and image captioning provide distinct forms of supervision. Supervision signals act as inductive biases that shape the structure of learned representations. While classification labels provide discriminative constraints, caption supervision encodes structured semantic descriptions that may encourage different forms of feature organization. While large-scale captioning models such as GIT~\citep{wang2022git}, BLIP~\citep{li2022blip}, BLIP-2~\citep{li2023blip}, LLaVA~\citep{liu2023visual}, and Kosmos-2~\citep{peng2023kosmos} illustrate the scalability of transformer-based multimodal architectures~\citep{vaswani2017attention,dosovitskiy2020image}, representational differences are often obscured by scale \citep{cornia2022explaining, elguendouze2023explainability} and evaluated primarily through downstream accuracy \citep{sariyildiz2020learning,hu2022scaling}. However, in small-data fine-grained settings, where uncertainty is higher and inductive biases play a stronger role, the choice of supervision modality may materially influence how models allocate attention to object parts and contextual cues.

This raises a fundamental question: does caption-based supervision induce measurably different visual representation structure compared to classification-based supervision, even when downstream predictive performance is similar? Answering this requires quantitative metrics and controlled comparisons that isolate supervision modality while holding architecture, optimization, and computational budget fixed.

Analyzing such differences requires systematic evaluation beyond accuracy. Attribution methods, including Integrated Gradients~\citep{sundararajan2017axiomatic}, SHAP~\citep{NIPS2017_7062}, LIME~\citep{ribeiro2016should}, Grad-CAM~\citep{selvaraju2017grad}, and attention rollout~\citep{abnar2020quantifying}, provide tools for probing model behavior. However, explanation quality is sensitive to the evaluation criterion. Recent work emphasizes the importance of multi-axis evaluation, assessing explainability across diverse objective dimensions to capture the complementary aspects of explanations~\citep{kazmierczak2025explainability,kibria2025decoding}. Such multi-criteria approaches are particularly valuable when comparing supervision regimes, where differences often emerge in representation alignment rather than task performance.

In this work, we investigate and quantify whether caption-based supervision transfers semantic structure which remains measurable after subsequent fine-grained classification tuning in a low-data setting. Using the CUB-200 dataset, which provides part-level annotations, we adopt a controlled two-stage protocol. In stage~1, two variants of an identical Vision Transformer (ViT) encoder~\citep{dosovitskiy2020image} are trained under different supervision: (i) classification via an attached linear head, and (ii) captioning using a frozen language decoder. In stage~2, both encoders are fine-tuned for fine-grained classification under matched optimization budgets. This design isolates the effect of supervision modality while controlling for architecture and compute.

\begin{figure*}[]
    \centering
    \includegraphics[width=.75\linewidth]{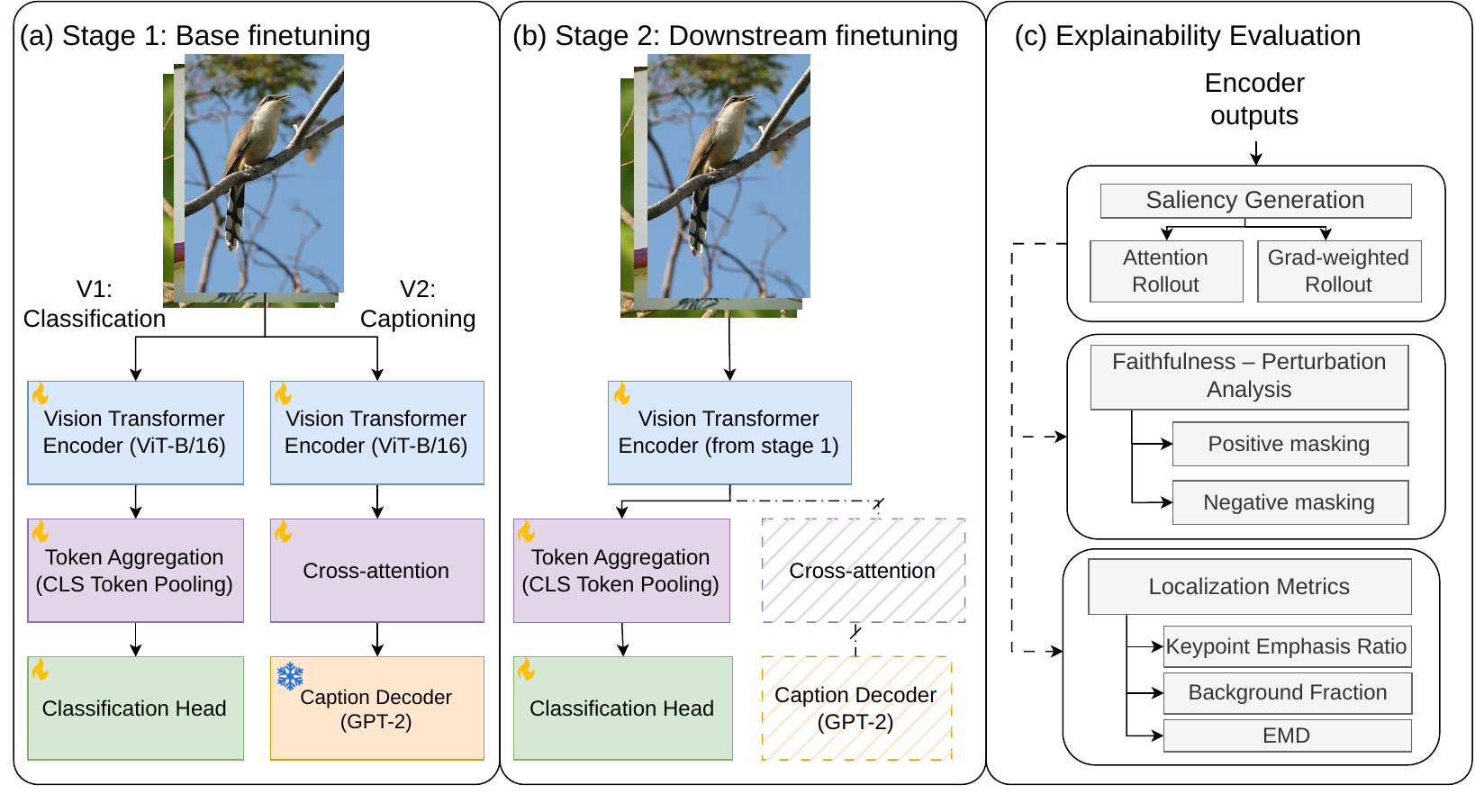}
    \caption{Overview of the proposed two-stage training and evaluation framework. (a) Stage 1 performs task-specific base tuning via either classification or captioning objectives. (b) Stage 2 applies controlled downstream finetuning for fine-grained classification using identical optimization budgets. (c) Explainability evaluation is conducted using multi-criteria metrics including perturbation-based faithfulness and localization measures.}
    \label{fig:block_arch}
\end{figure*}

To compare representation properties beyond predictive performance, we evaluate attribution behavior using attention rollout and gradient-augmented attention methods. We further introduce a structured evaluation protocol integrating five explainability metrics across two categories: localization, measured by keypoint emphasis, background fraction, and distributional alignment via Earth Mover's Distance (EMD); and faithfulness, measured by perturbation AUC under positive and negative masking. Leveraging CUB's part annotations enables objective quantification of semantic alignment. An overview of the proposed framework is presented in Figure~\ref{fig:block_arch}. We further assess robustness through two ablations: a cross-dataset evaluation on a second fine-grained benchmark and an encoder-scale ablation using a larger ViT from the same architectural family.

% Empirically, we find that caption-based pretraining yields consistent improvements in semantic alignment metrics, including stronger correspondence to annotated keypoints,  and reduced reliance on background regions. Although these improvements do not translate into higher downstream classification accuracy under fixed computational budgets, the representational gains are measurable and reproducible. In contrast, classification-tuning enables better exploitation of spurious cues. Taken together, our findings suggest that language supervision introduces a structural prior over visual representations, shaping object–context decomposition even when predictive risk remains largely unchanged.

% Empirically, caption-based pretraining does not improve downstream classification accuracy under fixed computational budgets. However, it yields consistent differences in attribution structure, including stronger correspondence to annotated keypoints, improved geometric alignment with keypoint-density maps, and reduced attribution to background regions. The ablations show that these directional trends persist under a second fine-grained dataset and a larger encoder scale. At the same time, classification-tuned encoders remain stronger under negative perturbation, indicating that the observed differences are not uniformly favorable to one supervision regimen across all criteria. Taken together, our findings demonstrate that the proposed framework can reveal supervision-dependent representation properties that are not captured by predictive accuracy alone.

Empirically, we find that caption-based pretraining produces consistent and reproducible differences in attribution structure, including stronger correspondence to annotated keypoints, improved geometric alignment with keypoint density maps, and reduced attribution to background regions. These trends persist across attribution methods, datasets, and encoder scales, suggesting that language supervision can shape visual representations in ways that are not reflected by classification accuracy alone. Although caption-based pretraining does not improve downstream accuracy under fixed computational budgets, the explainability-based analysis reveals a more general implication: supervision modality can affect the semantic organization of learned representations independently of predictive risk. Our results therefore position multi-criteria attribution evaluation as a useful tool for quantifying representation transfer beyond task performance.

\section{Related Work}

\paragraph{Language supervision in fine-grained recognition.}
Language supervision has been used to enrich visual representations in fine-grained settings such as CUB-200. \cite{hendricks2016generating} connected caption generation with grounded visual explanations, while \cite{reed2016learning} learned joint image–text embeddings that improved retrieval and zero-shot classification. Similarly, \cite{he2017fine} proposed a two stream model, separately trained for vision and language supervision for learning robust semantic representations. These works demonstrate that textual supervision can shape transferable features.

More recent multimodal representation learning studies including large-scale vision–language pretraining and contrastive objectives further show the potential of cross-modal transfers such as induction of semantically dense and localization-aware representations~\citep{radford2021learning, desai2021virtex, li2022grounded, el2023learning}. However, most studies evaluate such representations through downstream transfer performance, large-scale pretraining effectiveness, or task accuracy improvements, leaving open the question of how language supervision alters semantic alignment under controlled training budgets in small-data regimes.

\paragraph{Supervision modality and interpretability.}
Language-based supervision has also been studied in reasoning and multitask settings. The VCR benchmark~\citep{zellers2019recognition} highlights the role of language in grounding visual reasoning. \cite{parelli2023interpretable} distill reasoning supervision into visual models and report improvements in task-aligned faithfulness measures. Similarly, \cite{najjar2025can} show that auxiliary tasks can improve interpretability when strongly correlated with the primary objective. In contrast, we consider caption supervision applied only during pretraining and removed at evaluation time, enabling analysis of cross-modal transfer effects without explicit auxiliary objectives during classification.

\paragraph{Quantitative evaluation of representation alignment.}
For transformer-based models, attention-driven attribution methods such as attention rollout~\citep{abnar2020quantifying} and gradient-based localization~\citep{selvaraju2017grad, chefer2021generic} provide tools to probe learned representations. However, explainability is often evaluated using a single metric or qualitative inspection. Recent work argues that systematic evaluation of explainability should involve multi-dimensional analysis to account for their multi‑faceted and subjective nature ~\citep{nauta2023anecdotal,kibria2025decoding}. 

% Leveraging part annotations in CUB-200, we adopt a multi-criteria protocol that integrates keypoint coverage, background focus ratio, and perturbation-based faithfulness measures. This enables a controlled empirical study of how supervision modality influences semantic alignment independently of downstream accuracy.

Overall, prior work establishes that language supervision can improve transfer performance. Our contribution is to examine, under matched architectures and compute, whether caption-based pretraining induces measurable differences in representation alignment in low-data fine-grained settings.

\section{Methods}
This section describes the dataset, training protocols, explanation methods, and the proposed explainability evaluation framework.
% been observed to be robust against faithfulness violators, i.e., cases where the attribution assigns too much importance to regions not important for task objective. 

\subsection{Dataset}

We conduct experiments on the CUB-200-2011 dataset, which contains 11,788 images across 200 bird species and is widely used for fine-grained visual classification. In addition to class labels, the dataset provides object bounding boxes and 15 annotated part keypoints per image. These keypoints correspond to semantically meaningful bird parts (e.g., beak, wing, tail) and are commonly used to evaluate part-level localization. The dataset also includes binary attribute annotations and segmentation masks. Fine-grained visually descriptive captions without including the species or the background have been constructed in prior work for the dataset \citep{reed2016learning} and are used in this study for caption supervision. Further details regarding dataset annotations are provided in Appendix~\ref{sec:app_ds}.

\subsection{Models}

To evaluate the effect of supervision modality, we compare two training variants that differ only in the objective used during the first stage (Stage~1) of pretraining. Both variants share the same vision backbone: a Vision Transformer (ViT-B/16) initialized with ImageNet pretraining.

We note that ImageNet contains certain bird categories that may overlap semantically with CUB-200 classes. While exact image overlap is not present, this semantic proximity may influence representation transfer. However, this factor is not expected to confound our comparison because ImageNet initialization is held fixed across all experimental conditions, therefore, is used as a controlled common starting point to ensure stable optimization in the low-data regime; in contrast, training from scratch leads to substantially reduced performance and unstable convergence.

\paragraph{Classification-tuned variant.}
In the first variant, the encoder is trained with a standard classification objective by attaching a multi-layer perceptron (MLP) head to the [CLS] token representation.

\paragraph{Caption-tuned variant.}
In the second variant, caption supervision is introduced by attaching a pretrained GPT-2 decoder \citep{radford2019language} to the encoder through cross-attention layers. During Stage~1 training, most of the decoder parameters are frozen except the cross-attention weights, and the vision encoder is updated. This design isolates the effect of textual supervision on visual representations while preventing adaptation within the language model. Since the decoder is used only during Stage~1 training and primarily in the forward pass, with limited parameter updates, the additional computational overhead remains moderate.

In the second stage (Stage~2), both encoder variants are fine-tuned for downstream fine-grained classification under identical optimization settings. In this stage, a classification MLP head is trained on top of the encoder ([CLS] token representation), and no caption loss is applied. The overall two-stage architecture is illustrated in Figure~\ref{fig:block_arch} (a-b).

\subsection{Explanation Algorithms}

To analyze representation differences beyond predictive accuracy, we employ attribution methods tailored to transformer-based architectures.

\paragraph{Attention Rollout.}
We apply attention rollout \citep{abnar2020quantifying}, which aggregates attention weights across layers to estimate token-level influence. This method relies solely on forward-pass attention maps and produces class-agnostic attribution scores. Given that each image in CUB-200 contains a single labeled object, class-agnostic saliency remains informative for assessing spatial focus.

% Let $\bar{A}^{(l)} \in \mathbb{R}^{N \times N}$ denote the attention matrix of layer $l$, obtained by aggregating attention across heads. 
% To account for residual connections, we define the normalized layer attention by adding to it the identity matrix (Equation \ref{eq:rollout0}).
% \begin{equation}
% \tilde{A}^{(l)} = \mathrm{Norm}\!\left(\bar{A}^{(l)} + I \right),
% \label{eq:rollout0}
% \end{equation}
% where $\mathrm{Norm}(\cdot)$ denotes row-wise normalization.

% Attention rollout propagates attention through the network by matrix multiplication and token importance with respect to the $[\mathrm{CLS}]$ token is given by the first row (Equation \ref{eq:rollout}).
% \begin{equation}
% R = \prod_{l=1}^{L} \tilde{A}^{(l)}, 
% \qquad
% \mathbf{r} = R_{0,:}.
% \label{eq:rollout}
% \end{equation}

\paragraph{Gradient-weighted Attention Rollout.}
We additionally employ the transformer attribution method inspired from the work of \cite{chefer2021generic}, which combines attention weights with gradients of the target logit with respect to attention scores. This approach introduces class-specific information into the attribution and is conceptually related to gradient-based localization methods such as Grad-CAM \citep{selvaraju2017grad}. By incorporating gradient signals, the resulting saliency maps reflect the contribution of image regions to the predicted class.

The detailed mathematical formulation of both adapted algorithms is provided in Appendix~\ref{sec:app_math}.

Using both methods allows us to distinguish between general spatial attention patterns and class-conditional feature attribution, which is essential for evaluating representational differences induced by supervision modality.

\subsection{Explanation Evaluation}

Evaluating explanations requires multiple complementary criteria, as no single metric fully captures explanatory quality. We therefore assess explanations along two objective dimensions: \emph{faithfulness} and \emph{localization}. Faithfulness quantifies the extent to which saliency values reflect the true contribution of input regions to the model's prediction, while localization evaluates the spatial correspondence between saliency and annotated semantic structure. All metrics are computed identically across supervision variants.

% \subsubsection{Faithfulness: Positive and Negative Perturbation}
\subsubsection{Faithfulness: Positive and Negative Perturbation}

We adopt a perturbation-based protocol following \cite{chefer2021generic}, in which features are progressively masked according to their saliency ranking. The resulting change in prediction confidence is measured as the reflection of explanation faithfulness.

Let $S \in [0,1]^{H \times W}$ denote the normalized saliency map for an input image $I$. Since both attribution methods operate at patch level (ViT-B/16, 14$\times$14 grid), masking is performed over patches rather than individual pixels. Let $\tau \in \{0,5,10,\dots,100\}$ denote percentile thresholds over saliency values. We define two perturbation schemes:

\paragraph{Positive perturbation (most-salient first).}
Regions with saliency above threshold $\tau$ are masked, as in Equation \ref{eq:ppu}.

\begin{equation}
I^{+}_{\tau}(x,y) =
\begin{cases}
\mu, & S(x,y) \ge \tau \\
I(x,y), & \text{otherwise}
\end{cases}
\label{eq:ppu}
\end{equation}

\paragraph{Negative perturbation (least-salient first).}
Regions with saliency below threshold $\tau$ are masked, as in Equation \ref{eq:npu}.

\begin{equation}
I^{-}_{\tau}(x,y) =
\begin{cases}
I(x,y), & S(x,y) \ge \tau \\
\mu, & \text{otherwise}
\end{cases}
\label{eq:npu}
\end{equation}

Here, $\mu$ denotes the per-channel median pixel value computed over the training set. This baseline mitigates out-of-distribution artifacts that can arise from zero-masking.

For each $\tau$, we compute the predicted probability of the target class. Plotting confidence as a function of masking percentage yields perturbation curves. We summarize these curves using area under the curve (AUC) using $AUC_{I^{+}}$ and $AUC_{I^{-}}$, respectively for positive and negative perturbations. A faithful explanation yields a rapid decrease in confidence under positive perturbation resulting in a lower $AUC_{I^{+}}$ value whereas under negative perturbation, confidence should remain comparatively stable resulting in a higher $AUC_{I^{-}}$ value. The resulting $AUC_{I^{+}}$ and $AUC_{I^{-}}$ provide quantitative measures of sensitivity and robustness to feature removal.

A visual example of the progressive masking in both positive and negative directions is presented in Appendix \ref{sec:app_pnp}.

\subsubsection{Localization}

% Localization evaluates the spatial alignment between saliency and annotated semantic structure. 

Using the segmentation masks and keypoint annotations available in the dataset, we measure localization at complementary scales: object-level (fraction of background attribution), part-level (keypoint emphasis), and spatial-distribution alignment via EMD.

\paragraph{Background Fraction (Object-level localization).}

Let $B$ denote the set of background pixels derived from the segmentation mask. The background fraction measures the proportion of total saliency assigned to background regions as in Equation~\ref{eq:fbg}.

\begin{equation}
F_{bg} = 
\frac{\sum_{p \in B} S(p)}
{\sum_{p \in \text{image}} S(p)}.
\label{eq:fbg}
\end{equation}

Lower values indicate stronger saliency within object regions and reduced reliance on background context.

\paragraph{Keypoint Emphasis Ratio (Part-level localization).}

Fine-grained recognition often depends on localized discriminative parts, such as the beak, wing, or tail. To quantify how strongly a model emphasizes these regions, we introduce a new metric: the Keypoint Emphasis Ratio. Keypoint Emphasis Ratio measures the relative saliency assigned to annotated part locations compared to the remaining foreground region. Formally, let $\{C_k\}_{k=1}^{K}$ denote circular neighborhoods centered at the $K$ annotated keypoints. The radius $R$ is scaled with object size to ensure scale-adaptive neighborhoods calculated using Equation~\ref{eq:rad}.

\begin{equation}
R = \max\left(1, \left\lfloor \alpha \cdot \kappa \right\rfloor \right),
\label{eq:rad}
\end{equation}

where $\alpha \in (0,1]$ reflects normalized object area and $\kappa > 0$ controls spatial extent.

The average saliency within keypoint neighborhoods is

\[
\mathrm{KP}_{\mathrm{sal}} =
\frac{1}{K} \sum_{k=1}^{K}
\frac{1}{|C_k|}
\sum_{p \in C_k} S(p).
\]

Let $M$ denote the binary object mask. The mean saliency over object regions excluding keypoint neighborhoods is

\[
\mathrm{Obj}_{\mathrm{other\_sal}} =
\frac{1}{|M \setminus \cup_{k=1}^{K} C_k|}
\sum_{p \in M \setminus \cup_{k=1}^{K} C_k} S(p).
\]

We define the Keypoint Emphasis Ratio as in Equation~\ref{eq:rkp}.

\begin{equation}
R_{\mathrm{kp}} =
\frac{\mathrm{KP}_{\mathrm{sal}}}
{\mathrm{Obj}_{\mathrm{other\_sal}} + \epsilon},
\qquad \epsilon = 10^{-8}.
\label{eq:rkp}
\end{equation}

Values greater than one indicate preferential concentration of saliency around annotated discriminative parts relative to other object regions. A demonstration for example inputs on how the metric is calculated is presented in Appendix~\ref{sec:app_ker}.

\paragraph{Earth Mover's Distance (Spatial alignment).}

While $F_{\mathrm{bg}}$ and $R_{\mathrm{kp}}$ quantify saliency concentration within foreground and part regions, they do not account for spatial proximity. To measure geometric alignment between attribution maps and annotated discriminative parts, we compute the EMD between the attribution map and a Gaussianized keypoint density map.

% Given an attribution map $S$ and a Gaussianized keypoint density map $G$ constructed from annotated keypoints and masked by the object segmentation, spatial alignment is quantified as in Equation~\ref{eq:emd}.

% \begin{equation}
% D_{\mathrm{EMD}}
% =
% \mathrm{EMD}(\hat{S}, \hat{G}),
% \label{eq:emd}
% \end{equation}

Given a set of annotated keypoints ${(x_k,y_k)}_{k=1}^{K}$, we first construct a Gaussianized keypoint density map. Each keypoint contributes a Gaussian kernel centered at its annotated location, as in Equation~\ref{eq:gmap}.

\begin{equation}
G(p)
=
\sum_{k=1}^{K}
\exp \left(
-\frac{|p-(x_k,y_k)|_2^2}
{2\sigma^2}
\right),
\label{eq:gmap}
\end{equation}

where $\sigma$ controls the spatial extent of each keypoint neighborhood. The resulting density map is multiplied by the object segmentation mask to suppress background regions. Appendix~\ref{sec:app_emd} illustrates an example Gaussianized keypoint density map.

% \begin{figure}[t]
% \centering
% % \includegraphics[width=0.7\linewidth]{figures/gaussian_keypoint_map.pdf}
% \fbox{\parbox[c][4cm][c]{0.8\linewidth}{\centering Placeholder for Gaussianized keypoint density map}}
% \caption{Example Gaussianized keypoint density map generated from annotated object parts. Individual keypoints are represented by Gaussian kernels and masked using the object segmentation.}
% \label{fig}
% \end{figure}

The spatial alignment is then quantified from the attribution map $S$ and keypoint density map $G$, as in Equation~\ref{eq:emd}.

\begin{equation}
D_{\mathrm{EMD}}
=
\mathrm{EMD}(\hat{S}, \hat{G}),
\label{eq:emd}
\end{equation}

where $\hat{S}$ and $\hat{G}$ denote the normalized attribution and keypoint distributions, respectively. Lower values of $D_{\mathrm{EMD}}$ indicate stronger agreement between the spatial distribution of saliency and annotated discriminative parts. Unlike ratio-based metrics, EMD is geometry-aware and assigns smaller penalties to nearby localization errors than to distant ones, providing a complementary measure of spatial localization quality.

\section{Experiments}

This section details the training protocol, experimental setup, and associated hyperparameters.

\subsection{Training Setup}

Experiments were conducted on a single cluster node equipped with four NVIDIA V100 GPUs. Image captioning models were trained using the Huggingface API, enabling seamless multi-GPU training, while classification models were trained using PyTorch.

We use the CUB-200 dataset with 11,788 images, split into training, validation, and test subsets with proportions 50.9\%, 24.7\%, and 24.4\%, respectively. For caption supervision, we use the \cite{reed2016learning} caption extension, which provides ten natural-language descriptions per image, resulting in ten image--caption pairs for each image while preserving the same image-level splits used for classification. Captions are tokenized with the pretrained GPT-2 tokenizer, and images are processed using the standard ViT preprocessing pipeline~\citep{dosovitskiy2020image}.

% The total of 11,788 images in the dataset are split into training, validation, and test sets in proportions 50.9\%, 24.7\%, and 24.4\%, respectively. For caption supervision, we use the Reed et al. caption extension where each image has ten captions, often describing key visual attributes, increasing the effective training size for caption supervision tenfold. The same splits are used for both caption and classification training. Captions are tokenized using the pretrained GPT-2 tokenizer, and the ViT preprocessing method is used for the images \citep{dosovitskiy2020image}.

\subsection{Training Protocol}

\subsubsection{Stage 1: Base Encoder Tuning} 
During the first stage of training, two encoder variants are produced. Hyperparameters were chosen based on informed intuition and resulting experimentation without aiming for exhaustive optimization, as the optimal values are not crucial for the primary objectives of this study.

\begin{figure}[]
    \centering
    \begin{subfigure}[b]{0.65\linewidth}
        \centering
        \includegraphics[width=\linewidth]{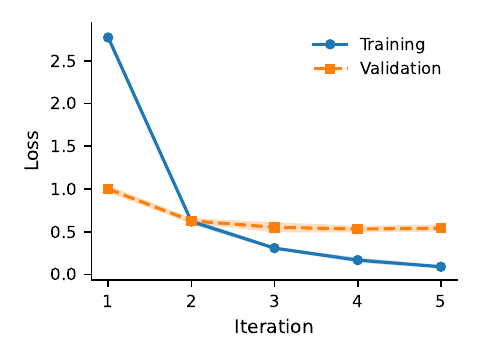}
        \caption{Training and Validation Loss}
        \label{fig:loss}
    \end{subfigure}
    \hfill
    \begin{subfigure}[b]{0.65\linewidth}
        \centering
        \includegraphics[width=\linewidth]{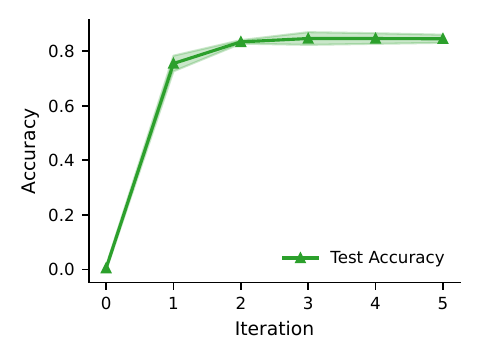}
        \caption{Test Accuracy}
        \label{fig:test_acc}
    \end{subfigure}
    \caption{Stage 1 training progress for the classification-tuned encoder. (a) training and validation loss; (b) test accuracy each averaged over three seeds with CI.}
    \label{fig:training_summary}
\end{figure}

\paragraph{Classification-tuned encoder.} The ImageNet-pretrained ViT is fine-tuned on CUB-200 for five iterations with a smaller learning rate for the encoder ($5\times10^{-5}$) and a larger rate for the classification head ($1\times10^{-3}$), enabling gradual adaptation of the backbone. Figure~\ref{fig:training_summary} shows the evolution of training and validation loss, and test accuracy averaged over three seeds with 95\% confidence intervals (CI). The final iteration achieves a mean test accuracy of 0.8454 ± 0.0148, with the highest mean accuracy observed at iteration 3.

\begin{figure}[!htb]
    \centering
    \includegraphics[width=\linewidth]{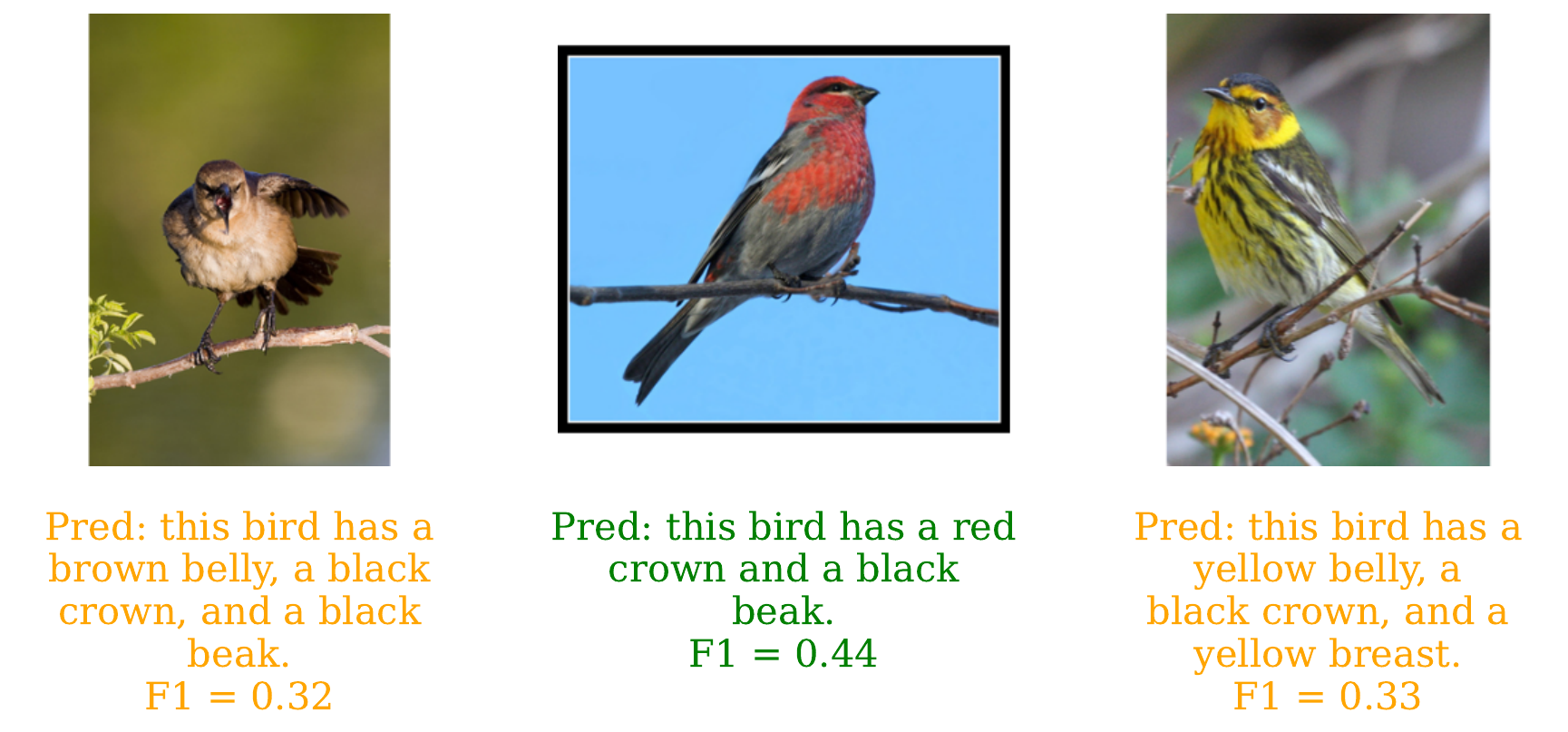}
    \caption{Qualitative captioning results under teacher forcing. Captions are color-coded based on maximum token-level F1 score against any of the 10 ground-truth captions (green: good match, orange: partial, red: poor). 
    % Partial fine-tuning produces moderate caption quality while preserving semantic grounding for downstream knowledge transfer.
    }
    \label{fig:cap_res}
\end{figure}

\paragraph{Caption-tuned encoder.} The image captioning pipeline is trained using a teacher-forcing approach. Here, only the encoder and the cross-attention weights of the decoder are fine-tuned, while the remaining decoder parameters remain frozen. Similar to the first variant, a smaller learning rate of $3\times10^{-5}$ is employed to train the encoder with a cross-entropy loss on the token outputs. Due to the limited diversity of captions in the dataset and the partial fine-tuning of the pipeline, the resulting captions exhibit only moderate quality, as illustrated in Figure~\ref{fig:cap_res}. Nevertheless, for the purposes of this study, the injection of external knowledge is prioritized over maximizing captioning accuracy.

\subsubsection{Stage 2: Downstream Classification Fine-Tuning} 
\begin{figure}[]
    \centering
    % ===== Row 1: Classification-tuned encoder =====
    \begin{subfigure}[b]{0.48\linewidth}
        \centering
        \includegraphics[width=\linewidth]{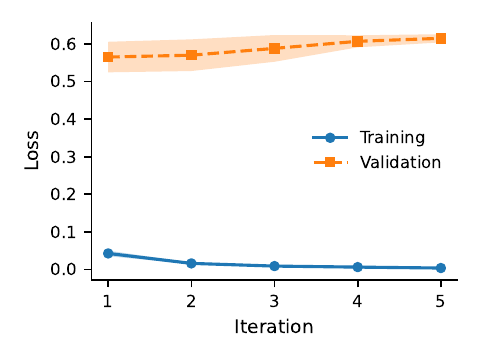}
        \caption{Classification encoder:\\Training and Validation Loss}
        \label{fig:class_loss}
    \end{subfigure}
    \hfill
    \begin{subfigure}[b]{0.48\linewidth}
        \centering
        \includegraphics[width=\linewidth]{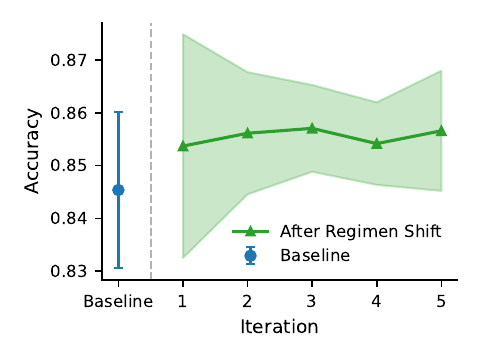}
        \caption{Classification encoder: Test Accuracy}
        \label{fig:class_acc}
    \end{subfigure}

    \vspace{0.5em}

    % ===== Row 2: Caption-tuned encoder =====
    \begin{subfigure}[b]{0.48\linewidth}
        \centering
        \includegraphics[width=\linewidth]{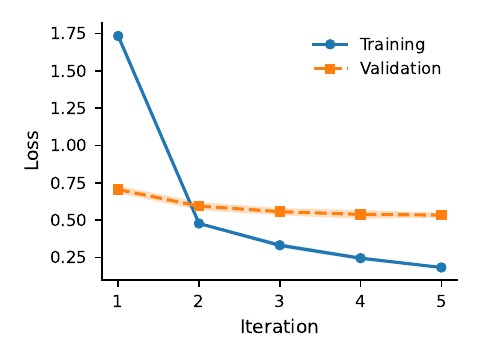}
        \caption{Caption encoder: Training and Validation Loss}
        \label{fig:cap_loss}
    \end{subfigure}
    \hfill
    \begin{subfigure}[b]{0.48\linewidth}
        \centering
        \includegraphics[width=\linewidth]{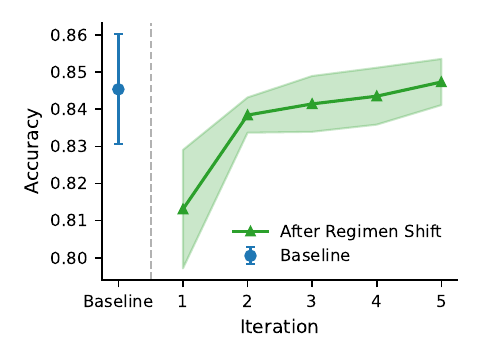}
        \caption{Caption encoder: Test Accuracy}
        \label{fig:cap_acc}
    \end{subfigure}

    \caption{Second-stage training progress over iterations averaged over three seeds with CI. Top row: classification-tuned encoder. Bottom row: caption-tuned encoder. Left: loss curves. Right: test accuracy with Stage~1 baseline.}
    \label{fig:second_stage_training}
\end{figure}

For each encoder variant, a classification head is added and fine-tuned for five iterations. A small learning rate ($1\times10^{-5}$) is used for the encoder, while the classification head uses a larger rate ($1\times10^{-3}$), primarily, to guide the caption-tuned encoder towards label alignment. Training summaries for both variants are shown in Figure~\ref{fig:second_stage_training}.

For the classification-tuned encoder, the training loss decreases gradually while validation loss remains stable (Figure~\ref{fig:class_loss}). Test accuracy rises steadily, with minor fluctuations due to the large head learning rate (Figure~\ref{fig:class_acc}). The caption-tuned encoder initially exhibits higher loss due to misalignment but stabilizes quickly, achieving comparable final accuracy. Final Stage~2 test accuracies (mean ± 95\% CI over five iterations) are summarized in Appendix~\ref{sec:app_acc}. Classification-tuned encoders achieve 0.8566 ± 0.0114, whereas caption-tuned encoders reach 0.8473 ± 0.0062, slightly lower under identical training budgets.

\section{Results}

We analyze the impact of the two training regimes beyond classification accuracy using explainability-oriented metrics computed over the test set across three random seeds. To assess whether supervision modality induces measurable differences in visual concept utilization, we report both quantitative analysis and qualitative trends.

\subsection{Perturbation-based Evaluation of Explanations}

We first evaluate explanation faithfulness using perturbation-based analysis. Following prior work, image regions are progressively masked according to saliency rankings while tracking the model's predicted probability for the ground-truth class. We consider positive and negative perturbations.

\begin{figure}[]
    \centering
    \begin{subfigure}[b]{0.9\linewidth}
        \centering
        \includegraphics[width=\linewidth]{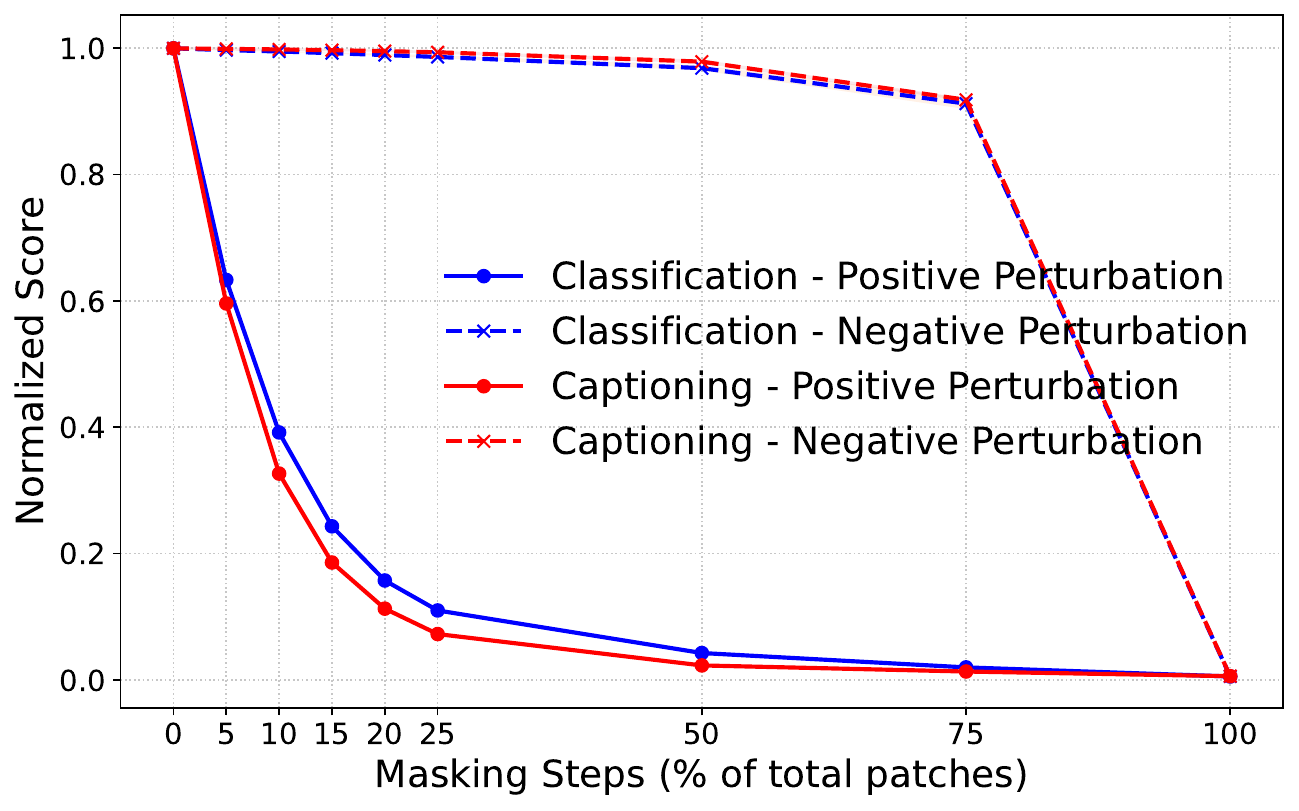}
        \caption{Perturbation Curves for Attention Rollout}
        \label{fig:ar_pnscores}
    \end{subfigure}
    \hfill
    \begin{subfigure}[b]{0.9\linewidth}
        \centering
        \includegraphics[width=\linewidth]{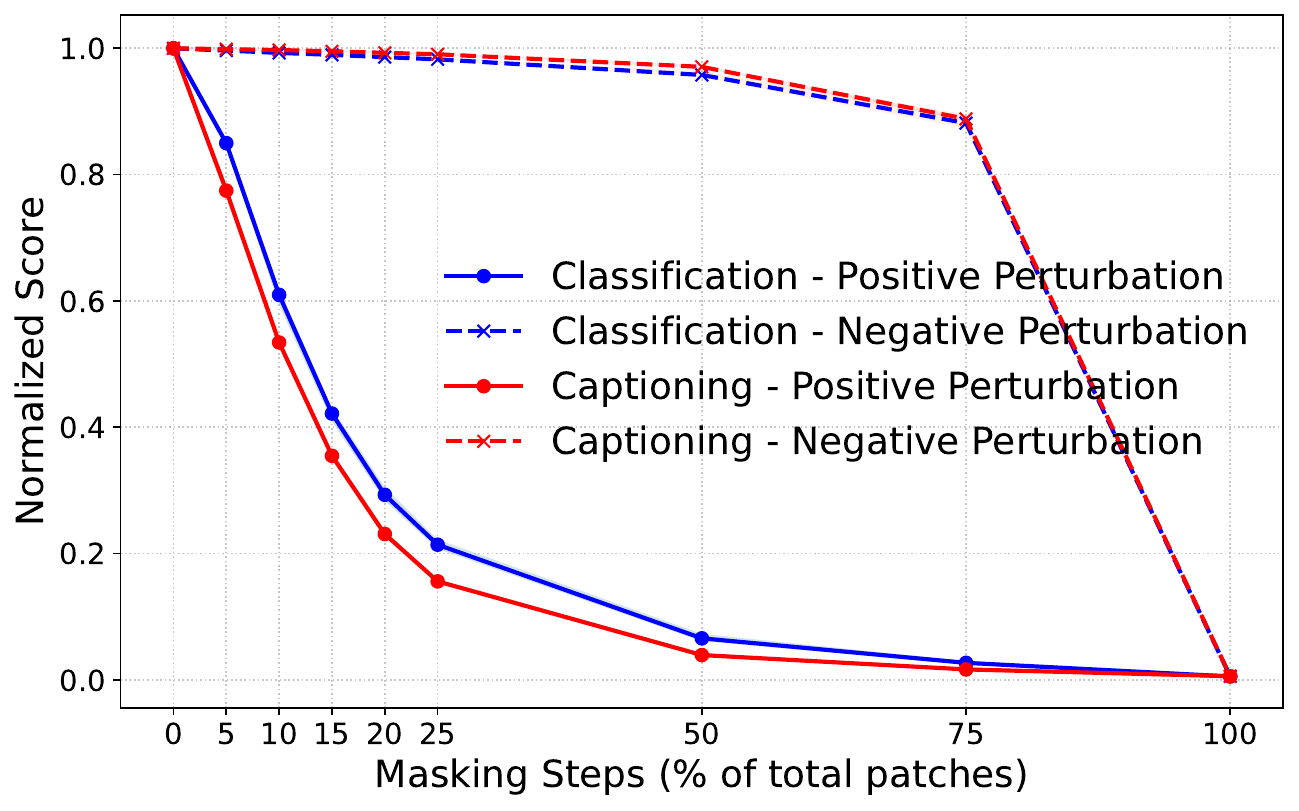}
        \caption{Perturbation Curves for Gradient-weighted Rollout}
        \label{fig:gar_pnscores}
    \end{subfigure}
    \caption{Positive and negative perturbation curves under progressive masking based on saliency maps from two explanation methods. \textit{Blue} curves correspond to the classification-tuned encoder and \textit{red} curves to the caption-tuned encoder. \textit{Solid} lines indicate positive perturbation (descending saliency masking), and \textit{dashed} lines indicate negative perturbation (ascending saliency masking). Curves are normalized by the unmasked prediction probability.}
    \label{fig:pnscores}
\end{figure}

Figure~\ref{fig:pnscores} shows the mean normalized prediction probability as a function of masking ratio for attention rollout and gradient-weighted attention explanations across classification- and caption-tuned encoders. The curves reflect explanation faithfulness in terms of sensitivity (positive perturbation) and robustness (negative perturbation).

Under negative perturbation, both encoders exhibit nearly similar stability across masking levels, indicating regions identified as minimally important can be removed without substantial degradation in predictive confidence. This suggests comparable robustness in discarding low-saliency regions, with the classification-tuned encoder showing a slight advantage.

In contrast, under positive perturbation, the caption-tuned encoder shows a consistently steeper decline in confidence as salient regions are progressively removed. This pattern appears across both attention-based and gradient-weighted explanations and persists throughout masking levels. The sharper degradation indicates that the caption-tuned model concentrates predictive mass on regions identified as salient, consistent with improved faithfulness.

Overall, these results suggest that caption supervision does not primarily increase reliance on contextual cues. Instead, it promotes stronger dependence on a smaller set of semantically meaningful regions, such that removing them results in a pronounced performance drop. While this increased sensitivity reflects improved alignment between saliency and model reliance, it does not imply enhanced robustness to distribution shift or adversarial perturbations, which remain open questions.

\subsection{Localization-based Evaluation of Explanations}

We next evaluate spatial alignment using three localization metrics: the Keypoint Emphasis Ratio ($R_{kp}$), the background fraction ($F_{bg}$) and EMD ($D_{EMD}$). These metrics compare saliency maps against ground-truth keypoint annotations and segmentation masks from complementary perspectives. For each seed, scores are averaged over the test set, yielding one aggregate value per training regime.

% The keypoint emphasis ratio $R_{kp}$ measures the relative concentration of saliency around annotated object keypoints. 
Higher values for the Keypoint Emphasis Ratio $R_{kp}$ indicate stronger alignment with discriminative object parts. The background fraction $F_{bg}$ quantifies the proportion of saliency assigned to background pixels; lower values correspond to more object-centric explanations. Given the fine-grained nature of the dataset, reduced background attribution is a meaningful indicator of improved localization. $D_{EMD}$ captures the spatial alignment between the attribution distribution and Gaussianized keypoint density maps. Thus, lower $D_{EMD}$ values indicate better localization and closer correspondence between model attributions and keypoint density.

\begin{figure*}[]
    \centering
    \begin{subfigure}[b]{.8\linewidth}
        \centering
        \includegraphics[width=\linewidth]{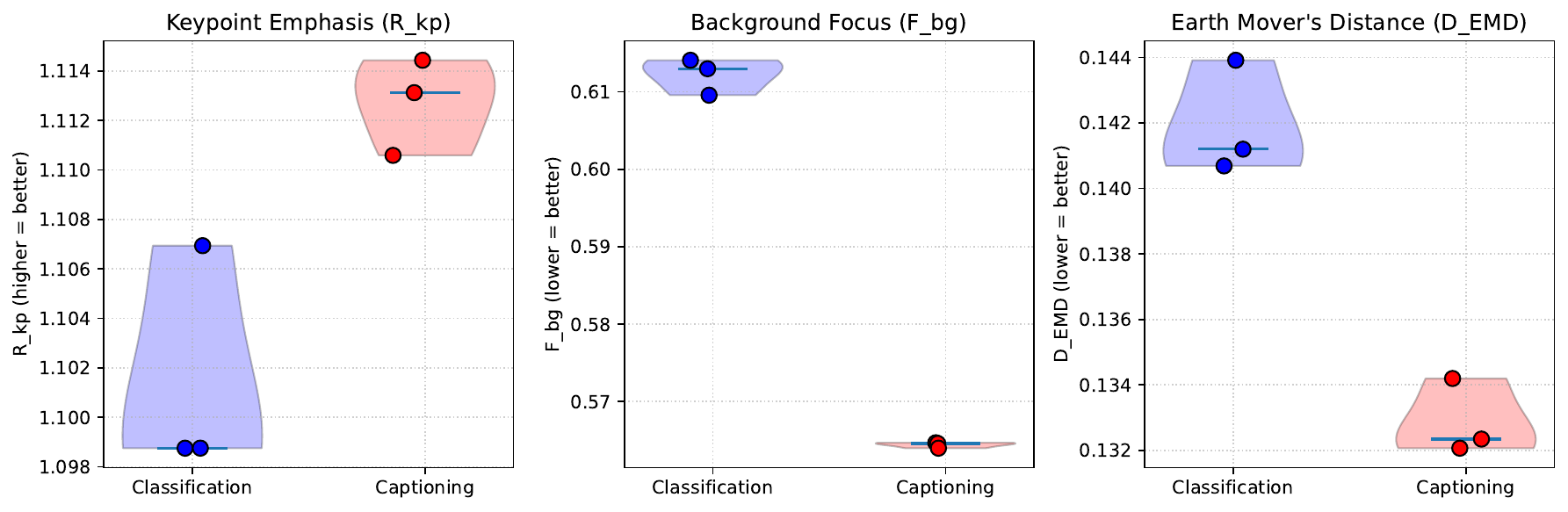}
        \caption{Distribution of Localization Metrics for Attention Rollout}
        \label{fig:ar_loc}
    \end{subfigure}
    \hfill
    \begin{subfigure}[b]{.8\linewidth}
        \centering
        \includegraphics[width=\linewidth]{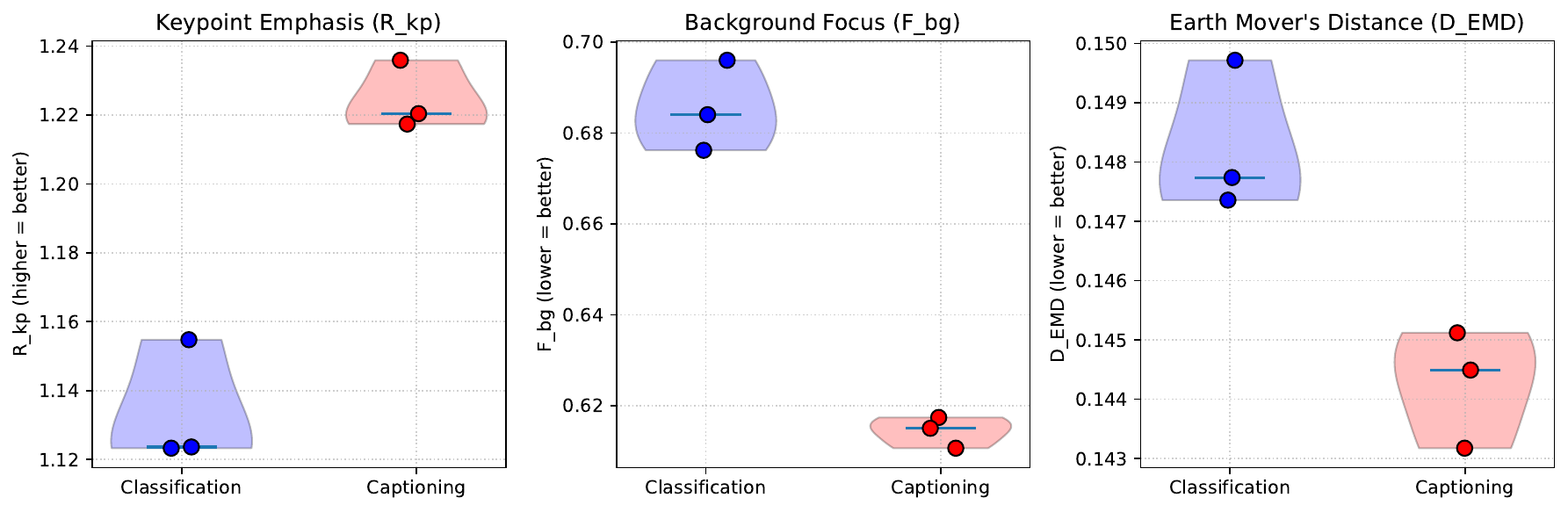}
        \caption{Distribution of Localization Metrics for Gradient-weighted Rollout}
        \label{fig:gar_loc}
    \end{subfigure}
    \caption{Violin plots of localization metrics across three random seeds for classification- and caption-tuned encoders. Results are shown for (a) attention rollout and (b) gradient-weighted rollout. 
    % The keypoint emphasis ratio ($R_{kp}$, higher is better) measures alignment with annotated keypoints, while the background fraction ($F_{bg}$, lower is better) quantifies saliency assigned to background regions.
    \textit{Blue} violins correspond to the classification-tuned encoder and \textit{red} violins to the caption-tuned encoder. Points denote seed-wise averages and the central horizontal line indicates the median.}
    \label{fig:violin}
\end{figure*}

Figure~\ref{fig:violin} presents violin plots of $R_{kp}$, $F_{bg}$ and $D_{EMD}$ across three seeds for both training regimes and explanation methods.

Across both explanation methods, caption tuning consistently yields higher $R_{kp}$ values with reduced inter-seed variability compared to classification tuning. This indicates stronger and more stable alignment with discriminative object parts. This improvement is also maintained across different radius values (see Appendix~\ref{sec:radius_ablation}). Similarly, caption-tuned encoders produce uniformly lower $F_{bg}$ values, reflecting reduced saliency allocation to background regions. 

While the improvement in $D_{EMD}$ is smaller than for the ratio-based metrics, caption-tuned models consistently obtain lower values across both attribution methods and all seeds, indicating closer alignment between the saliency and target spatial distributions.

Collectively, these results indicate that caption supervision systematically biases representations toward foreground-centric and semantically grounded explanations. The consistency of these effects across two attribution methods and multiple seeds suggests that the observed improvements arise from the training objective rather than artifacts of a particular saliency algorithm.

\subsection{Statistical Analysis of the Metrics}

All five metrics were found to be non-normally distributed using the Shapiro–Wilk test. We therefore apply the Wilcoxon signed-rank test to compare paired per-image metric values between classification- and caption-tuned encoders. Effect sizes are reported as paired Cohen's $d$. For each seed, metrics were computed over the full test set; pooled paired samples across three seeds yield $n=8{,}646$ comparisons per metric. Statistical results for both explanation methods are summarized in Table~\ref{tab:stats}.

\begin{table*}[]
\caption{Comparison between Classification- and Captioning-tuned encoder training regimens using Wilcoxon signed-rank tests ($n=8646$ paired samples per metric). Effect sizes reported as paired Cohen's $d$. For the five metrics, expected directions are: $AUC_{I^{+}}$ ↓; $AUC_{I^{-}}$ ↑; $F_{bg}$ ↓; $R_{kp}$ ↑; $D_{EMD}$ ↓.}
\centering
\small
\begin{tabular}{>{\raggedright\arraybackslash}p{2.4cm}lcccccc}
\toprule
Saliency Method & Metric & Regimen & Mean & Std & 95\% CI & p-value & Cohen's d \\
\midrule

\multirow{10}{=}{Grad-Weighted Attention}
& \multirow{2}{*}{$AUC_{I^{+}}$}
    & Classification & 0.1762 & 0.1430 & [0.1733, 0.1793] & \multirow{2}{*}{$5.33\times10^{-307}$} & \multirow{2}{*}{0.388} \\
&   & Captioning & \textbf{0.1328} & 0.1074 & [0.1305, 0.1351] &  &  \\\cmidrule(lr){2-8}

& \multirow{2}{*}{$AUC_{I^{-}}$}
    & Classification & \textbf{0.7716} & 0.1844 & [0.7675, 0.7754] & \multirow{2}{*}{$<10^{-308}$} & \multirow{2}{*}{0.405} \\
&   & Captioning & 0.7048 & 0.2032 & [0.7004, 0.7091] &  &  \\
\cmidrule(lr){2-8}

& \multirow{2}{*}{$F_{bg}$}
    & Classification & 0.6854 & 0.1072 & [0.6831, 0.6877] & \multirow{2}{*}{$<10^{-308}$} & \multirow{2}{*}{0.741} \\
&   & Captioning & \textbf{0.6144} & 0.1115 & [0.6120, 0.6168] &  &  \\
\cmidrule(lr){2-8}

& \multirow{2}{*}{$R_{kp}$}
    & Classification & 1.1339 & 0.3318 & [1.1271, 1.1411] & \multirow{2}{*}{$1.32\times10^{-96}$} & \multirow{2}{*}{-0.227} \\
&   & Captioning & \textbf{1.2246} & 0.4299 & [1.2156, 1.2338] &  &  \\
\cmidrule(lr){2-8}

& \multirow{2}{*}{$D_{EMD}$}
    & Classification & 0.1483 & 0.0304 & [0.1476, 0.1489] & \multirow{2}{*}{$2.38\times10^{-39}$} & \multirow{2}{*}{0.1286} \\
&   & Captioning & \textbf{0.1443} & 0.0329 & [0.1436, 0.1449] &  &  \\

\midrule

\multirow{10}{=}{Attention Rollout}
& \multirow{2}{*}{$AUC_{I^{+}}$}
    & Classification & 0.1199 & 0.1167 & [0.1174, 0.1224] & \multirow{2}{*}{$<10^{-308}$} & \multirow{2}{*}{0.410} \\
&   & Captioning & \textbf{0.0898} & 0.0806 & [0.0880, 0.0915] &  &  \\
\cmidrule(lr){2-8}

& \multirow{2}{*}{$AUC_{I^{-}}$}
    & Classification & \textbf{0.7822} & 0.1767 & [0.7782, 0.7859] & \multirow{2}{*}{$<10^{-308}$} & \multirow{2}{*}{0.434} \\
&   & Captioning & 0.7135 & 0.1987 & [0.7092, 0.7176] &  &  \\
\cmidrule(lr){2-8}

& \multirow{2}{*}{$F_{bg}$}
    & Classification & 0.6122 & 0.1184 & [0.6096, 0.6147] & \multirow{2}{*}{$<10^{-308}$} & \multirow{2}{*}{1.768} \\
&   & Captioning & \textbf{0.5644} & 0.1225 & [0.5617, 0.5670] &  &  \\
\cmidrule(lr){2-8}

& \multirow{2}{*}{$R_{kp}$}
    & Classification & 1.1015 & 0.1925 & [1.0976, 1.1056] & \multirow{2}{*}{$2.56\times10^{-145}$} & \multirow{2}{*}{-0.220} \\
&   & Captioning & \textbf{1.1127} & 0.1896 & [1.1089, 1.1168] &  &  \\
\cmidrule(lr){2-8}

& \multirow{2}{*}{$D_{EMD}$}
    & Classification & 0.1419 & 0.0249 & [0.1414, 0.1424] & \multirow{2}{*}{$<10^{-308}$} & \multirow{2}{*}{0.8882} \\
&   & Captioning & \textbf{0.1328} & 0.0247 & [0.1323, 0.1333] &  &  \\

\bottomrule
\end{tabular}
\label{tab:stats}
\end{table*}

All metrics show statistically significant paired differences ($p \ll 0.001$). For perturbation-based faithfulness, the results show a consistent trade-off. Caption tuning consistently yields lower values across both saliency methods for $AUC_{I^{+}}$, with small-to-moderate effect sizes ($d=0.388, 0.410$), reflecting stronger confidence degradation when salient regions are removed. In contrast, classification tuning achieves higher $AUC_{I^{-}}$, indicating greater stability when low-saliency regions are removed. Thus, caption tuning produces stronger sensitivity to removal of salient evidence, while classification tuning remains more robust to removal of low-ranked regions.

For localization, caption tuning produces lower $F_{bg}$ with moderate-to-large effect sizes, indicating reduced background attribution. Gains in $R_{kp}$ for caption-tuning are smaller (small effect sizes) but consistent across the saliency methods, indicating stronger alignment with annotated keypoints. The same trend is reflected by $D_{EMD}$, where caption tuning improves geometric alignment with keypoint locations, with a small effect for Grad-Weighted Rollout and a larger effect for Attention Rollout.

Together, the statistical results confirm systematic differences in representation structure induced by supervision modality that are not captured by classification accuracy alone.

\begin{figure*}[]
    \centering
    \includegraphics[width=.85\linewidth]{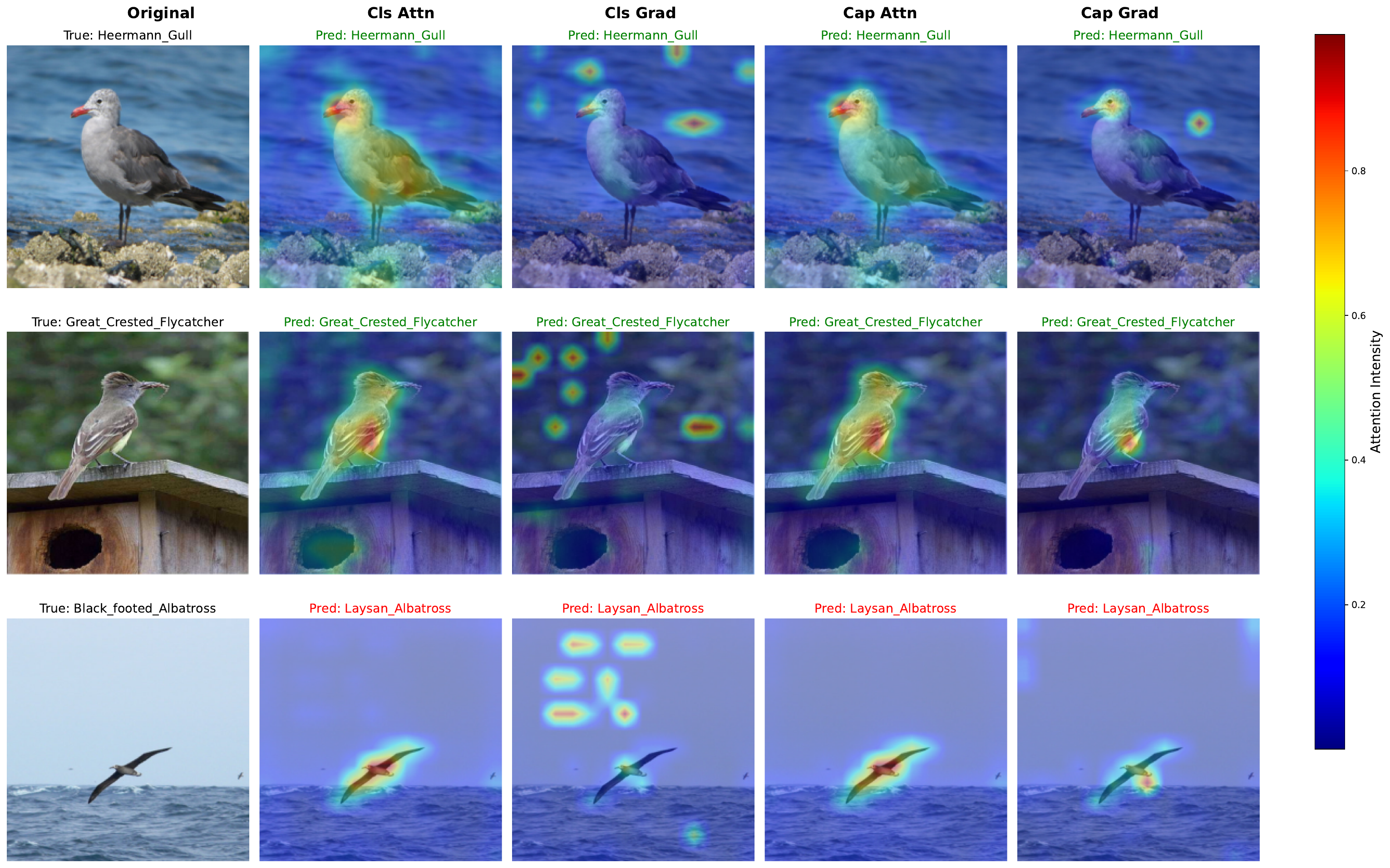}
    \caption{
        Saliency map visualizations for classification- and caption-tuned models. 
        Each row corresponds to one test image. 
        Column 1 shows the original image with the true label. 
        Columns 2–5 show saliency overlays: classification-tuned and caption-tuned encoders both under attention rollout and gradient-weighted rollout. 
        Predicted labels are shown above each heatmap (green = correct, red = incorrect). 
        A shared colorbar indicates saliency intensity.
    }
    \label{fig:attention_visualizations}
\end{figure*}

\subsection{Qualitative Comparison}

To complement quantitative analysis, we present representative test images with predicted labels and overlaid saliency maps for both training regimes and explanation methods in Figure~\ref{fig:attention_visualizations}.

Across examples, gradient-weighted rollout exhibits higher-frequency artifacts, particularly in background regions, likely due to the sensitivity of gradients and upsampling from patch to pixel resolution. Attention rollout produces smoother, more spatially coherent maps.

Comparing training regimes, differences are subtle but consistent with quantitative trends. Caption-tuned models show reduced background activation and more spatially concentrated attributions, particularly under gradient-weighted rollout. This pattern appears for both correct and incorrect predictions, suggesting a structural shift in how evidence is distributed rather than a byproduct of accuracy differences.

While qualitative maps support the statistical findings, they should be interpreted cautiously given the non-causal nature of saliency methods. Conclusions rely primarily on systematic quantitative evaluation.

\section{Ablation Studies}

We conduct two ablation studies to examine whether the trends observed on CUB-200 are robust to changes in dataset and model capacity. First, we evaluate the same protocol on the Oxford-IIIT Pet dataset, which provides a second fine-grained recognition setting with foreground segmentation masks. Second, we repeat the CUB-200 experiment using a larger ViT-L/16 encoder to test whether the attribution trends persist beyond a single encoder scale.

% To evaluate whether the observed effects are specific to CUB-200 or reflect a broader pattern in fine-grained recognition, we conduct an additional ablation on the Oxford-IIIT Pet dataset. This experiment tests the robustness of the same two-stage training and explainability protocol under a different category structure and annotation regime.

\subsection{Results on the Oxford-IIIT Pet Dataset}

The Oxford-IIIT Pet dataset contains 37 fine-grained pet categories, comprising 12 cat breeds and 25 dog breeds, with 7,349 images in total. In contrast to CUB-200, the dataset does not provide part-level keypoint annotations, but includes pixel-level segmentation masks separating the foreground pet from the background. We therefore evaluate background reliance using the background saliency fraction, $F_{bg}$, together with the perturbation-based faithfulness metrics. For the captioning objective, we use an augmented version of the dataset in which image captions describing the foreground pet and surrounding scene are generated using BLIP-2~\citep{li2023blip}. We use an 80--10--10 train-validation-test split of the original dataset.

We apply the same two-stage protocol as in the main CUB-200 experiments: task-specific encoder tuning with either classification or captioning supervision, followed by controlled downstream classification fine-tuning under matched optimization budgets. We reduced the second-stage finetuning schedule relative to CUB-200 (2 epochs instead of 5), as the smaller dataset reached stable validation performance with fewer updates, however, comparison between supervision modalities remains controlled. Table~\ref{tab:cub_ablation} reports results over three random seeds using Wilcoxon signed-rank tests on paired test examples.

\begin{table*}[]
\caption{Ablation on Oxford-IIIT Pet comparing classification- and caption-tuned encoders.
Each cell reports mean $\pm$ standard deviation over paired test samples. 
$\Delta$ denotes Captioning $-$ Classification. 
Arrows indicate the preferred direction.}
\centering
\small
\setlength{\tabcolsep}{5pt}
\begin{tabular}{llccccc}
\toprule
 & 
& Classification
& Captioning
& 
& 
&  \\
Saliency Method & Metric
& (Best accuracy:
& (Best accuracy:
& $\Delta$
&  $p$
& $d$ \\
 & 
& $93.83\% \pm 0.0098$)
& $92.37\% \pm 0.0071$)
& 
& 
& \\
\midrule

\multirow{3}{*}{Grad-Weighted Attention}
& $AUC_{I^{+}}$ $\downarrow$
& $0.3747 \pm 0.2092$
& $\mathbf{0.3118 \pm 0.1829}$
& $-0.0630$
& $2.28{\times}10^{-99}$
& $0.436$ \\

& $AUC_{I^{-}}$ $\uparrow$
& $\mathbf{0.7744 \pm 0.1570}$
& $0.7571 \pm 0.1601$
& $-0.0173$
& $1.83{\times}10^{-31}$
& $0.124$ \\

& $F_{bg}$ $\downarrow$
& $0.5278 \pm 0.1600$
& $\mathbf{0.4793 \pm 0.1515}$
& $-0.0485$
& $1.16{\times}10^{-93}$
& $0.559$ \\
\midrule

\multirow{3}{*}{Attention Rollout}
& $AUC_{I^{+}}$ $\downarrow$
& $0.3249 \pm 0.2184$
& $\mathbf{0.2785 \pm 0.1836}$
& $-0.0464$
& $6.65{\times}10^{-71}$
& $0.383$ \\

& $AUC_{I^{-}}$ $\uparrow$
& $\mathbf{0.7976 \pm 0.1491}$
& $0.7721 \pm 0.1520$
& $-0.0255$
& $1.04{\times}10^{-81}$
& $0.208$ \\

& $F_{bg}$ $\downarrow$
& $0.4217 \pm 0.1471$
& $\mathbf{0.3764 \pm 0.1473}$
& $-0.0453$
& $<10^{-308}$
& $0.986$ \\

\bottomrule
\end{tabular}
\label{tab:cub_ablation}
\end{table*}

The Oxford-IIIT Pet ablation shows the same qualitative pattern as the main CUB-200 analysis. Caption tuning consistently reduces $AUC_{I^{+}}$ across both saliency methods, indicating sharper degradation when the most salient regions are removed. It also reduces $F_{bg}$, suggesting lower background attribution and stronger concentration on the foreground object. These effects are statistically significant for both explanation methods, with small-to-moderate effects for perturbation faithfulness and moderate-to-large effects for background reliance.

% Across both saliency methods, all metrics exhibit statistically significant paired differences ($p \ll 0.001$). The direction of effects is consistent with those observed on CUB-200. Caption tuning yields lower $AUC_{I^{+}}$ with small-to-moderate effect sizes ($d=0.383$–$0.436$), indicating stronger degradation when salient regions are removed, and thus improved faithfulness. Similarly, $F_{bg}$ is consistently reduced, with moderate-to-large effect sizes (up to $d=0.986$), reflecting decreased reliance on background regions.

As in CUB-200, classification-tuned models retain higher $AUC_{I^{-}}$, indicating weaker sensitivity to removal of least salient regions. While this effect remains statistically significant, its magnitude is comparatively smaller.

Overall, despite differences in dataset structure and annotation granularity, the qualitative trends are preserved. This consistency supports the conclusion that supervision modality induces systematic and reproducible differences in representation structure, rather than dataset-specific artifacts.

\subsection{Encoder-Scale Ablation with ViT-L/16}

To assess whether the observed trends are specific to the ViT-B/16 backbone, we conduct an encoder-scale ablation using an ImageNet-pretrained ViT-L/16. This preserves the same architectural family and attribution pipeline while evaluating the framework under a higher-capacity encoder. The two-stage training protocol and matched downstream tuning setup are kept consistent with the main experiments, with the optimization budget adjusted for the larger model. 

Table~\ref{tab:single_seed_ablation} shows that the ViT-L/16 results preserve the same directional trends observed in the main experiments. Across both Grad-Weighted Rollout and Attention Rollout, caption tuning improves the preferred direction for $AUC_{I^{+}}$, $F_{bg}$, $R_{kp}$, and $D_{EMD}$, while classification tuning remains stronger for $AUC_{I^{-}}$. Although the magnitude of the differences is generally smaller than with ViT-B/16, particularly for Attention Rollout, the consistency across metrics and attribution methods supports the robustness of the observed supervision-dependent attribution patterns.

\begin{table*}[]
\caption{Encoder-scale ablation on CUB-200 using ViT-L/16.
Each cell reports mean $\pm$ standard deviation over $n=2882$ paired test samples.
$\Delta$ denotes Captioning $-$ Classification.
Arrows indicate the preferred direction.}
\centering
\small
\setlength{\tabcolsep}{4pt}
\begin{tabular}{llccccc}
\toprule
 & 
& Classification
& Captioning
& 
& 
&  \\
Saliency Method & Metric
& (Best accuracy:
& (Best accuracy:
& $\Delta$
&  $p$
& $d$ \\
 & 
& $87.05\%$)
& $84.52\%$)
& 
& 
& \\
\midrule

\multirow{5}{*}{Grad-Weighted Rollout}
& $AUC_{I^{+}}$ $\downarrow$
& $0.1430 \pm 0.1222$
& $\mathbf{0.1272 \pm 0.0944}$
& $-0.0158$
& $2.89{\times}10^{-11}$
& $0.175$ \\

& $AUC_{I^{-}}$ $\uparrow$
& $\mathbf{0.7688 \pm 0.1919}$
& $0.7124 \pm 0.2149$
& $-0.0563$
& $1.64{\times}10^{-144}$
& $0.313$ \\

& $F_{bg}$ $\downarrow$
& $0.6528 \pm 0.0925$
& $\mathbf{0.5942 \pm 0.1080}$
& $-0.0585$
& $8.19{\times}10^{-212}$
& $0.674$ \\

& $R_{kp}$ $\uparrow$
& $1.2035 \pm 0.3420$
& $\mathbf{1.2700 \pm 0.4101}$
& $0.0665$
& $9.52{\times}10^{-21}$
& $-0.187$ \\

& $D_{EMD}$ $\downarrow$
& $0.1358 \pm 0.0293$
& $\mathbf{0.1286 \pm 0.0293}$
& $-0.0072$
& $1.95{\times}10^{-33}$
& $0.224$ \\

\midrule

\multirow{5}{*}{Attention Rollout}
& $AUC_{I^{+}}$ $\downarrow$
& $0.1813 \pm 0.1411$
& $\mathbf{0.1567 \pm 0.1258}$
& $-0.0246$
& $1.40{\times}10^{-45}$
& $0.254$ \\

& $AUC_{I^{-}}$ $\uparrow$
& $\mathbf{0.7512 \pm 0.1997}$
& $0.6983 \pm 0.2196$
& $-0.0530$
& $2.48{\times}10^{-113}$
& $0.292$ \\

& $F_{bg}$ $\downarrow$
& $0.6867 \pm 0.1272$
& $\mathbf{0.6803 \pm 0.1271}$
& $-0.0064$
& $6.26{\times}10^{-67}$
& $0.296$ \\

& $R_{kp}$ $\uparrow$
& $1.0464 \pm 0.1509$
& $\mathbf{1.0686 \pm 0.1849}$
& $0.0222$
& $1.51{\times}10^{-51}$
& $-0.301$ \\

& $D_{EMD}$ $\downarrow$
& $0.1653 \pm 0.0281$
& $\mathbf{0.1628 \pm 0.0280}$
& $-0.0026$
& $3.23{\times}10^{-104}$
& $0.390$ \\

\bottomrule
\end{tabular}
\label{tab:single_seed_ablation}
\end{table*}

\section{Discussion and Conclusion}

In this work, we present a controlled empirical framework for analyzing multitask transferability from image captioning to fine-grained classification through explainability-based quantification of learned visual representations. Using CUB-200 as the primary benchmark, we compare classification- and caption-tuned Vision Transformer encoders under a two-stage protocol: task-specific base tuning followed by shared downstream classification finetuning under matched optimization budgets. This design enables attribution-level comparison of supervision regimes while controlling for architecture, downstream task, and compute.

The main contribution of this study is a multi-criteria evaluation framework that links representation transfer to measurable explanation properties. Rather than relying only on classification accuracy, we combine faithfulness-based metrics, through positive and negative perturbation AUC, with localization-based metrics, including keypoint emphasis, background attribution, and spatial-distribution alignment. This provides a structured way to quantify how different supervision modalities organize visual evidence after transfer.

Across the main CUB-200 experiments, caption-based pretraining produces consistent advantages on most attribution-based metrics. Caption-tuned encoders show stronger correspondence to annotated keypoints, improved geometric alignment with keypoint density maps, reduced attribution to background regions, and greater sensitivity to removal of highly salient regions. Although these improvements do not translate into higher downstream classification accuracy under the fixed training budget, they reveal systematic differences in representation structure that predictive performance alone does not capture. This suggests that language supervision can shape how visual evidence is organized, even when the final task risk remains comparable. In contrast, classification tuning produces slightly greater stability under the removal of low-saliency regions.

The ablation studies further support the utility of the framework. On Oxford-IIIT Pet, where only foreground segmentation is available, the same protocol identifies consistent supervision-dependent trends using background attribution and perturbation metrics. The ViT-L/16 encoder-scale ablation preserves the directional patterns observed with ViT-B/16, indicating that the analysis is not tied to a single encoder capacity. Together, these experiments show that the proposed framework can be applied across datasets, annotation granularities, and model scales to quantify how training objectives affect learned representations.

Overall, the findings highlight the value of explainability-based evaluation for studying representation transfer beyond accuracy. Caption-based pretraining is not superior in downstream classification accuracy, but it induces more semantically structured attribution behavior across most evaluated criteria. This distinction is important for settings where the organization of evidence, localization of task-relevant regions, and reduced background reliance matter alongside predictive performance. Future work can extend this framework to larger-scale multimodal objectives, including contrastive vision--language learning, additional fine-grained and domain-specific datasets to further characterize how supervision modality shapes visual representations.

\backmatter

% \bmhead{Supplementary information}

% If your article has accompanying supplementary file/s please state so here. 

% Authors reporting data from electrophoretic gels and blots should supply the full unprocessed scans for key as part of their Supplementary information. This may be requested by the editorial team/s if it is missing.

% Please refer to Journal-level guidance for any specific requirements.

% \bmhead{Acknowledgements}

% Acknowledgements are not compulsory. Where included they should be brief. Grant or contribution numbers may be acknowledged.

% Please refer to Journal-level guidance for any specific requirements.

% \begin{itemize}
% \item Funding
% \item Conflict of interest/Competing interests (check journal-specific guidelines for which heading to use)
% \item Ethics approval and consent to participate
% \item Consent for publication
% \item 
% % \item Materials availability
% % \item Code availability 
% \item Author contribution
% \end{itemize}

\section*{Data availability} All datasets used in this study are publicly available. The CUB-200-2011 dataset is available from its official distribution. The natural language captions for CUB-200-2011 are those provided by \cite{reed2016learning}. The Oxford-IIIT Pet dataset is publicly available, and the enriched captions used in this work are obtained from the Visual Layer Oxford-IIIT Pet VL Enriched dataset available at \url{https://huggingface.co/datasets/visual-layer/oxford-iiit-pet-vl-enriched/}. No new datasets were generated during this study.

\section*{Author contribution} \textbf{M.R.K.:} Conceptualization, Methodology, Investigation, Software, Formal analysis, Visualization, Writing—original draft. \textbf{S.L.:} Conceptualization, Methodology, Supervision, Writing—review and editing. All authors read and approved the final manuscript.

\section*{Declarations}
The authors declare no conflict of interest.

\section*{Funding}
The authors received no financial support for the research, and/or authorship of this article.

\section*{Code Availability}
The source code, trained models, and scripts required to reproduce the experiments will be made publicly available upon publication.

\section*{Ethics Approval} 
Not applicable.

\bibliography{bib}
\begin{appendices}

% \section{Section title of first appendix}\label{secA1}

\section{Dataset Description} \label{sec:app_ds}
\begin{figure}[b]
    \centering
    \begin{subfigure}[b]{0.8\linewidth}
        \centering
        \includegraphics[width=\linewidth]{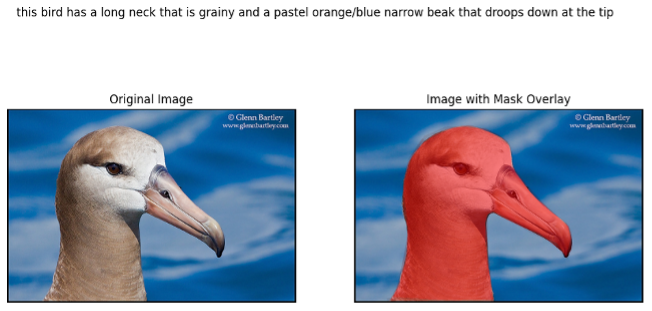}
        \caption{Original input image (left) and the same image with the object mask overlay (right), along with one example caption (top).}
    \end{subfigure}
    \hfill
    \begin{subfigure}[b]{0.8\linewidth}
        \centering
        \includegraphics[width=\linewidth]{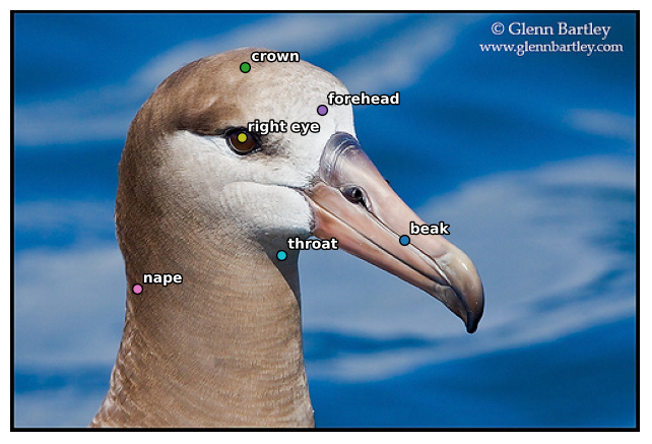}
        \caption{Annotated part names with their corresponding keypoint locations for the example image.}
    \end{subfigure}
    \caption{Illustration of CUB-200-2011 dataset annotations, including object masks and part locations.}
    \label{fig:dataset}
\end{figure}

An example of the dataset annotations is shown in Figure~\ref{fig:dataset}. The dataset provides 15 annotated object parts (keypoints): \textit{back, beak, belly, breast, crown, forehead, left eye, left leg, left wing, nape, right eye, right leg, right wing, tail}, and \textit{throat}.

\section{Explanation Algorithm Formulations} \label{sec:app_math}
Here, we present the formulation of the adaptation of the explanation algorithms applied in this work.
\paragraph{Attention Rollout.}
Let $\bar{A}^{(l)} \in \mathbb{R}^{N \times N}$ denote the attention matrix of layer $l$, obtained by aggregating attention across heads. 
We incorporate residual connections and normalize row-wise as in Equation~\ref{eq:att_l}.
\begin{equation}
\tilde{A}^{(l)} = \mathrm{Norm}\!\left(\bar{A}^{(l)} + I \right),
\label{eq:att_l}
\end{equation}
where $\mathrm{Norm}(\cdot)$ denotes row normalization.

Attention is propagated through the network as in Equation~\ref{eq:roll}.
\begin{equation}
R = \prod_{l=1}^{L} \tilde{A}^{(l)},
\qquad
\mathbf{r} = R_{0,:}.
\label{eq:roll}
\end{equation}

\paragraph{Gradient-Weighted Attention Rollout.}
Let $A^{(l,h)}$ denote the attention matrix of head $h$ in layer $l$, and let 
$G^{(l,h)}$ be the gradient of the target output $y$ with respect to that attention matrix. 
We define a scalar importance weight per head as $\alpha_h^{(l)}$ and construct the gradient-weighted layer attention as in Equation~\ref{eq:ahat}.

\begin{equation}
\hat{A}^{(l)} = \sum_{h} \alpha_h^{(l)} A^{(l,h)} .
\label{eq:ahat}
\end{equation}
where,
\[
G^{(l,h)} = \frac{\partial y}{\partial A^{(l,h)}} 
\]
\[
\alpha_h^{(l)} = \mathrm{Mean}\!\left(G^{(l,h)}\right)
\]

The normalized layer attention is then calculated using Equation~\ref{eq:awave}.
\begin{equation}
\tilde{A}^{(l)} = \mathrm{Norm}\!\left(\hat{A}^{(l)} + I \right),
\label{eq:awave}
\end{equation}
and rollout is computed identically as in Equation~\ref{eq:roll}.
% \begin{equation}
% R = \prod_{l=1}^{L} \tilde{A}^{(l)},
% \qquad
% \mathbf{r} = R_{0,:}.
% \end{equation}

\section{Positive and Negative Perturbation Demonstration}\label{sec:app_pnp}
\begin{figure}[]
    \centering
    \subfloat[Saliency map (attention rollout).]{
        \includegraphics[width=0.3\textwidth]{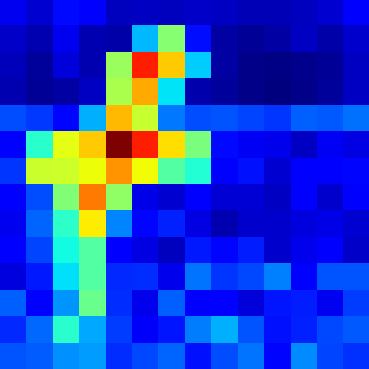}
    }
    \hspace{0.02\textwidth}
    \subfloat[Positive perturbation (most-salient regions removed first).]{
        \includegraphics[width=0.3\textwidth]{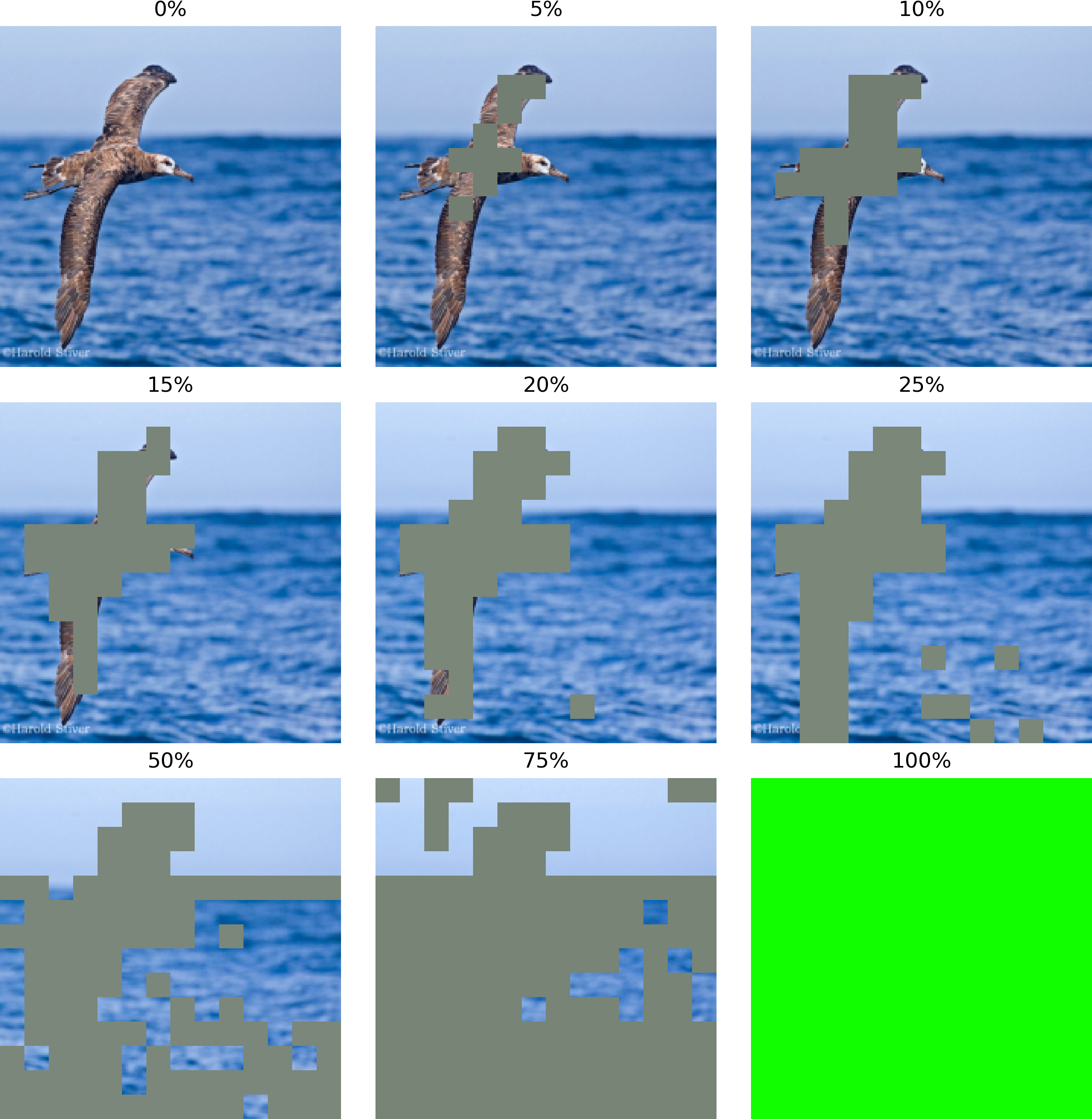}
    }
    \hspace{0.02\textwidth}
    \subfloat[Negative perturbation (least-salient regions removed first).]{
        \includegraphics[width=0.3\textwidth]{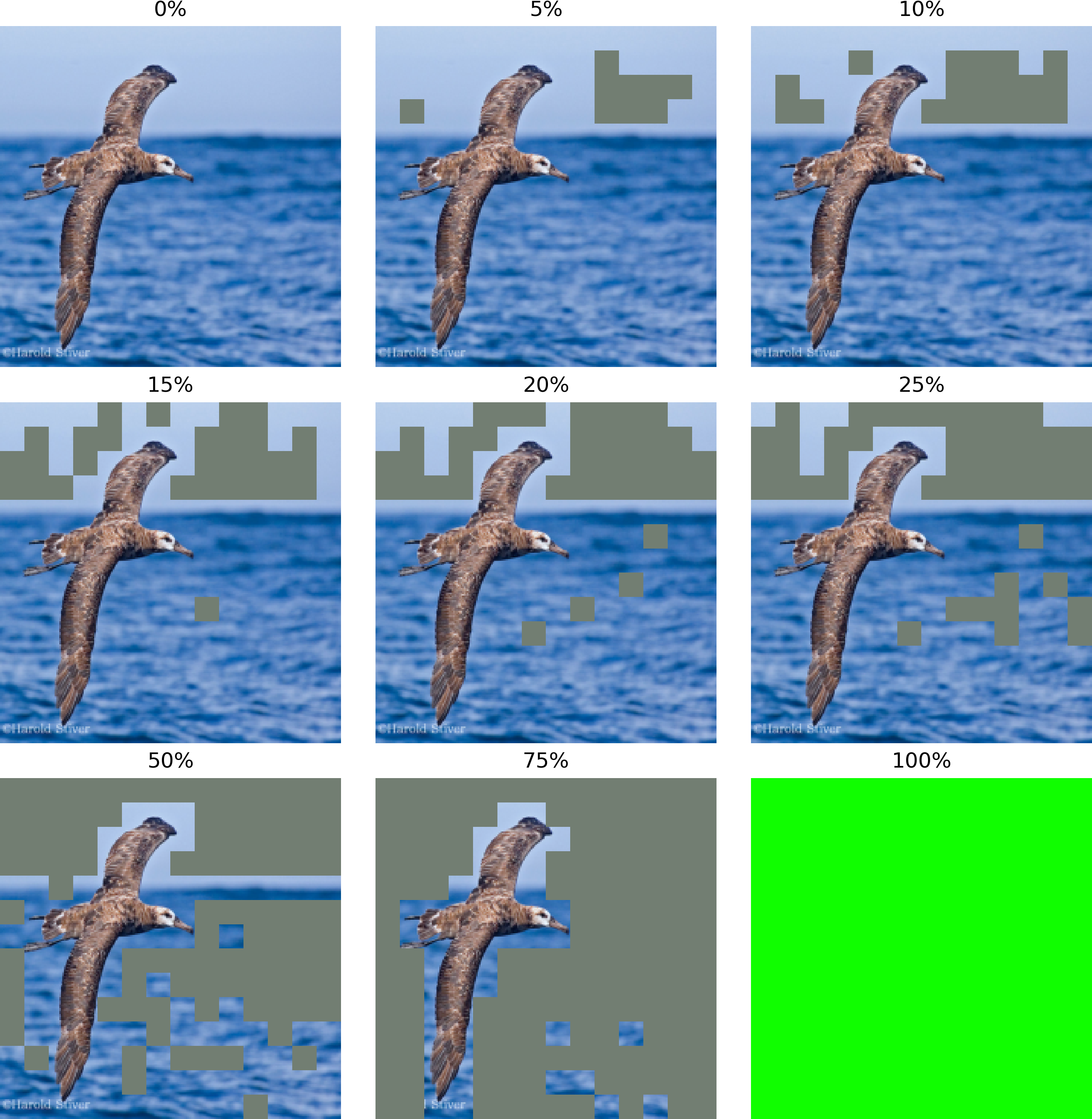}
    }
    \caption{Saliency-based perturbation protocol. Positive perturbation removes the most-salient regions first, producing a rapid drop in model confidence if the explanation is faithful. Negative perturbation removes the least-salient regions first, preserving task-relevant evidence and resulting in comparatively stable confidence.}
    \label{fig:saliency_perturbation}
\end{figure}

An example of how an attribution map derived for a sample test input from the attention rollout method is used to measure positive and negative perturbation progressively is presented in Figure~\ref{fig:saliency_perturbation}.

\section{Keypoint Emphasis Ratio Demonstration}\label{sec:app_ker}
\begin{figure*}[]
    \centering
    \includegraphics[width=.9\linewidth, trim={0 500pt 0 0}, clip]{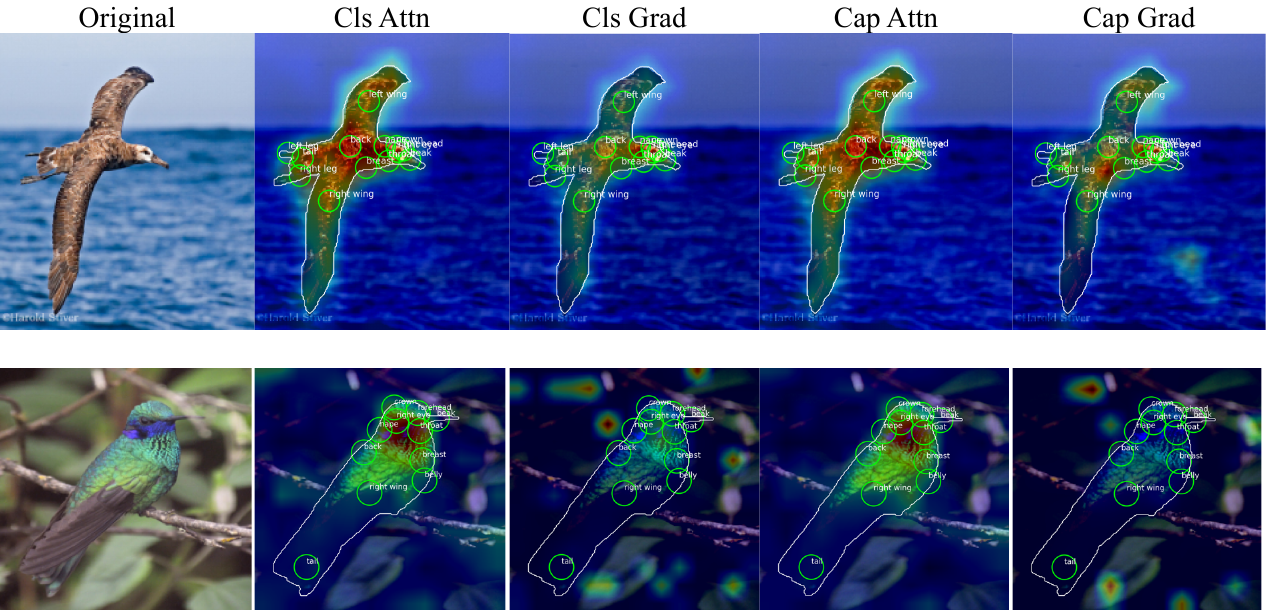}
    \caption{Illustration of the Keypoint Emphasis Ratio ($R_{kp}$), which quantifies the relative concentration of saliency around annotated object keypoints. Each row shows one example. Column 1 presents the original image with its ground-truth label. Columns 2–5 display saliency overlays for classification- and caption-tuned encoders under attention rollout and gradient-weighted rollout. \textit{Green} circles denote annotated keypoints with a scale-adaptive radius (spatial extent $0.25$) used to compute $R_{kp}$. The \textit{white} contour defines the object boundary. \textit{Red} indicates high attribution intensity and \textit{blue} indicates low attribution intensity.}
    \label{fig:ker_demo}
\end{figure*}

Figure~\ref{fig:ker_demo} illustrates how the Keypoint Emphasis Ratio is measured for two representative test inputs across the two training regimes and both attribution methods.

\section{Earth Mover's Distance Demonstration\label{sec:app_emd}}
\begin{figure*}[]
\centering
\includegraphics[width=0.7\linewidth]{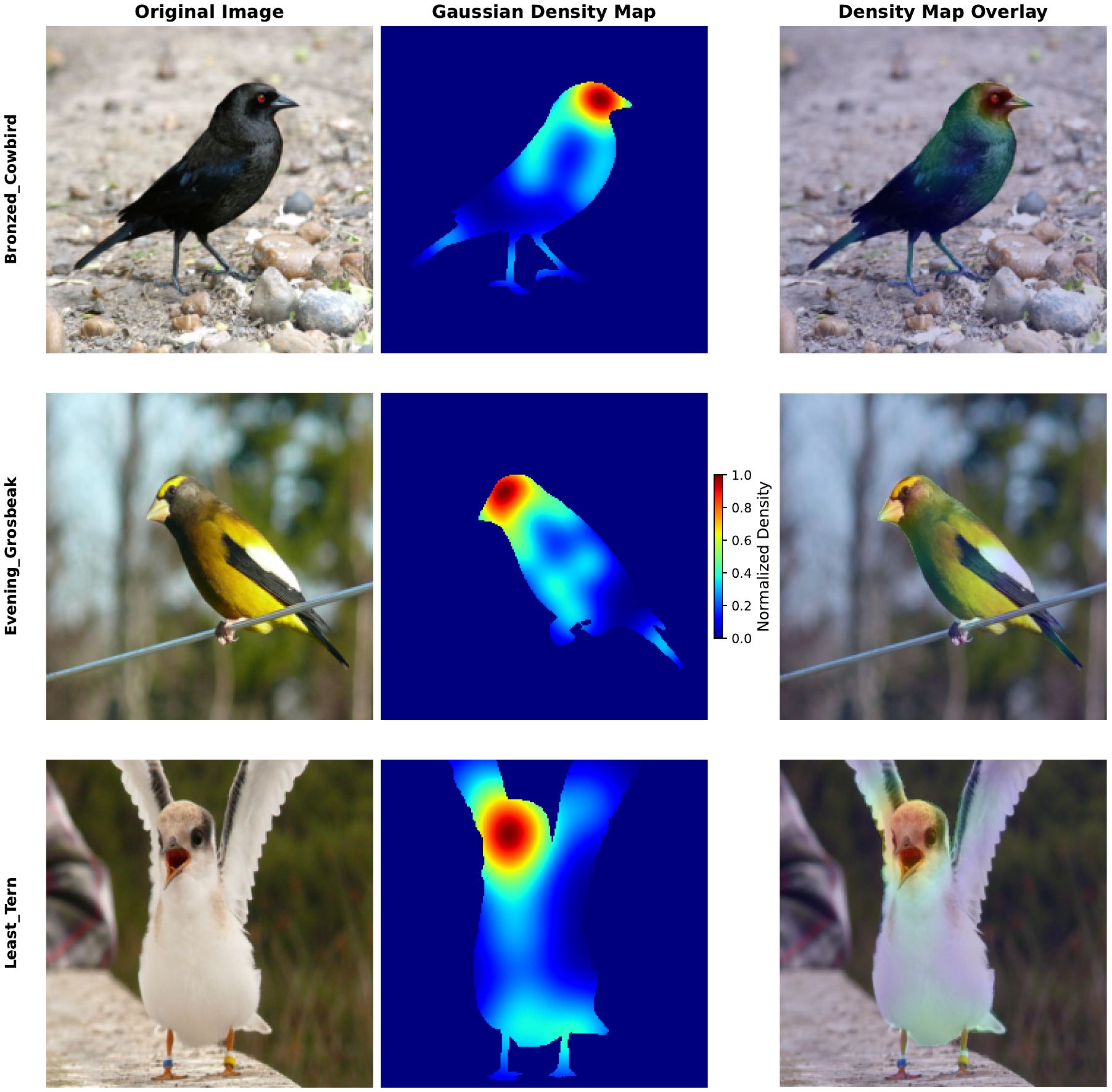}
\caption{Example Gaussianized keypoint density map generated from annotated object parts. Individual keypoints are represented by Gaussian kernels and masked using the object segmentation.}
\label{fig:emd}
\end{figure*}

Example of how keypoints are used to create a Gaussianized density maps is presented in Figure~\ref{fig:emd}.

\section{Two Stage Training Summary}\label{sec:app_acc}

Table~\ref{tab:acc} summarizes the training behavior of the two supervision regimes by reporting test-set classification accuracy on CUB-200 across different training iterations. This comparison provides an overview of how predictive performance evolves during training for the classification-based and caption-based settings under the same evaluation protocol.

\begin{table*}[]
\centering
\caption{Two-stage training results. Stage~2 follows Stage~1 (5 Iterations).}
\label{tab:acc}
\begin{tabular}{c p{5cm} c p{5cm}}
\hline
\textbf{Stage} & \textbf{Training Regimen} & \textbf{Iteration} & \textbf{Test Accuracy (mean ± 95\% CI)} \\
\hline
\multirow{6}{*}{Stage 1} 
& \multirow{6}{=}{V1: Classification-base\\training (used for Stage 2A)} 
   & 0 & 0.0049 ± 0.0082 \\
&   & 1 & 0.7547 ± 0.0286 \\
&   & 2 & 0.8339 ± 0.0065 \\
&   & 3 & 0.8463 ± 0.0232 \\
&   & 4 & 0.8462 ± 0.0194 \\
&   & 5 & \textbf{0.8454 ± 0.0148} \\
\cmidrule(lr){2-4}
& V2: Captioning-base training (used for Stage 2B) & Post training & 0.0367 ± 0.0254 \\
\hline
\multirow{5}{*}{Stage 2} 
& \multirow{5}{=}{A: Final tuning\\of classification-tuned encoder\\(from Stage 1 V1)} 
        & 1 & 0.8537 ± 0.0212 \\
        && 2 & 0.8561 ± 0.0116 \\
        && 3 & 0.8570 ± 0.0082 \\
        && 4 & 0.8542 ± 0.0078 \\
        && 5 & \textbf{0.8566 ± 0.0114} \\
\cmidrule(lr){2-4}
\multirow{5}{*}{} 
& \multirow{5}{=}{B: Final tuning\\of caption-tuned encoder (from Stage 1 V2)} 
        & 1 & 0.8131 ± 0.0159 \\
        && 2 & 0.8384 ± 0.0047 \\
        && 3 & 0.8414 ± 0.0075 \\
        && 4 & 0.8435 ± 0.0077 \\
        && 5 & \textbf{0.8473 ± 0.0062} \\
\hline
\end{tabular}
\end{table*}

\section{Ablating Radius for Keypoint Emphasis Calculation}\label{sec:radius_ablation}
We report the Keypoint Emphasis Ratio, $R_{kp}$, across spatial extension ratios $\kappa \in \{0.15, 0.25, 0.35, 0.50\}$. Results for a single random seed, comparing both training regimes and saliency methods, are shown in Table~\ref{tab:radius_ablation}.

\begin{table*}[]
\centering
\caption{Radius Ablation across Saliency Algorithms and Training Regimes. 
Mean Keypoint Emphasis Ratio, $R_{kp}$, reported for four scale factors.}
\label{tab:radius_ablation}

\begin{tabular}{llcccc}
\toprule
\textbf{Saliency Algorithm} & \textbf{Training Regime} 
& \textbf{0.15} & \textbf{0.25 (default)} & \textbf{0.35} & \textbf{0.50} \\
\midrule

\multirow{2}{*}{Attention Rollout} 
& Captioning 
& \textbf{1.083620} & \textbf{1.114433} & \textbf{1.153625} & \textbf{1.225190} \\
\cmidrule(lr){2-6}

& Classification 
& 1.070167 & 1.098750 & 1.134251 & 1.199433 \\
\hline
\multirow{2}{*}{Grad-Weighted Rollout}
& Captioning 
& \textbf{1.173308} & \textbf{1.235900} & \textbf{1.315662} & \textbf{1.463961} \\
\cmidrule(lr){2-6}

& Classification 
& 1.109976 & 1.154779 & 1.215932 & 1.337401 \\

\bottomrule
\end{tabular}
\end{table*}

Across all spatial extensions, the captioning regime consistently achieves higher $R_{kp}$ values, indicating stronger alignment with semantically meaningful keypoints. This trend holds for both small and large neighborhoods around annotations and across saliency methods, reinforcing the main localization findings. Gradient-weighted rollout yields overall higher $R_{kp}$ scores compared to attention rollout, though the relative advantage of caption supervision remains consistent within each method.

%%=============================================%%
%% For submissions to Nature Portfolio Journals %%
%% please use the heading ``Extended Data''.   %%
%%=============================================%%

%%=============================================================%%
%% Sample for another appendix section			       %%
%%=============================================================%%

%% \section{Example of another appendix section}\label{secA2}%
%% Appendices may be used for helpful, supporting or essential material that would otherwise 
%% clutter, break up or be distracting to the text. Appendices can consist of sections, figures, 
%% tables and equations etc.

\end{appendices}

%%===========================================================================================%%
%% If you are submitting to one of the Nature Portfolio journals, using the eJP submission   %%
%% system, please include the references within the manuscript file itself. You may do this  %%
%% by copying the reference list from your .bbl file, paste it into the main manuscript .tex %%
%% file, and delete the associated \verb+\bibliography+ commands.                            %%
%%===========================================================================================%%

% \bibliography{bib}% common bib file
% %% if required, the content of .bbl file can be included here once bbl is generated
% %%\input sn-article.bbl

\end{document}